%% file: WCNC2025.tex
\documentclass[conference]{IEEEtran}
\IEEEoverridecommandlockouts

\usepackage{cite}
\usepackage{amsmath,amssymb,amsfonts}
\usepackage{algorithmic}
\usepackage{graphicx}
\usepackage{textcomp}
\usepackage{xcolor}
\usepackage{paralist}
\usepackage{eurosym}
\usepackage{flushend}
\usepackage{url}
\usepackage[utf8]{inputenc}
\usepackage[colorlinks=true]{hyperref}
\usepackage{soul}
\usepackage{caption}
\usepackage[list=true]{subcaption}
\usepackage{booktabs}
\usepackage{tabu}
\usepackage{wrapfig}
\usepackage{arydshln}
\usepackage[normalem]{ulem}
\usepackage{multirow}
\usepackage{siunitx}

\usepackage{alphalph}

\def\BibTeX{{\rm B\kern-.05em{\sc i\kern-.025em b}\kern-.08em
    T\kern-.1667em\lower.7ex\hbox{E}\kern-.125emX}}
    
\begin{document}

\title{Federated Learning with MMD-based Early Stopping for Adaptive GNSS Interference Classification\\
\thanks{This work has been carried out within the DARCII project, funding code 50NA2401, sponsored by the German Federal Ministry for Economic Affairs and Climate Action (BMWK) and supported by the German Aerospace Center (DLR), the Bundesnetzagentur (BNetzA), and the Federal Agency for Cartography and Geodesy (BKG).}}

\input{00authors.tex}

\maketitle

\input{00abstract.tex}
\begin{IEEEkeywords}
  Federated Learning, Few-Shot Learning, Representation Learning, Maximum Mean Discrepancy, GNSS Interference Classification, Sensor Orchestration
\end{IEEEkeywords}

\input{01introduction.tex}
\input{02related_work.tex}
\input{03method.tex}
\input{04experiments.tex}
\input{05evaluation.tex}
\input{06conclusion.tex}

\bibliography{WCNC2025}
\bibliographystyle{IEEEtran}

\end{document}

%% file: 00authors.tex
\author{\IEEEauthorblockN{Nishant S. Gaikwad, Lucas Heublein, Nisha L. Raichur, Tobias Feigl, Christopher Mutschler, Felix Ott}
  \IEEEauthorblockA{Fraunhofer Institute for Integrated Circuits IIS, Nürnberg, Germany}
  \IEEEauthorblockA{{\{nishant.shankar.gaikwad, lucas.heublein, nisha.lakshmana.raichur, tobias.feigl, christopher.mutschler,} \IEEEauthorblockA{\underline{felix.ott}\}@iis.fraunhofer.de}}}

%% file: 00abstract.tex
\begin{abstract}
Federated learning (FL) enables multiple devices to collaboratively train a global model while maintaining data on local servers. Each device trains the model on its local server and shares only the model updates (i.e., gradient weights) during the aggregation step. A significant challenge in FL is managing the feature distribution of novel and unbalanced data across devices. In this paper, we propose an FL approach using few-shot learning and aggregation of the model weights on a global server. We introduce a dynamic early stopping method to balance out-of-distribution classes based on representation learning, specifically utilizing the maximum mean discrepancy of feature embeddings between local and global models. An exemplary application of FL is to orchestrate machine learning models along highways for interference classification based on snapshots from global navigation satellite system (GNSS) receivers. Extensive experiments on four GNSS datasets from two real-world highways and controlled environments demonstrate that our FL method surpasses state-of-the-art techniques in adapting to both novel interference classes and multipath scenarios.\\
\href{https://gitlab.cc-asp.fraunhofer.de/darcy_gnss/federated_learning}{https://gitlab.cc-asp.fraunhofer.de/darcy\_gnss/federated\_learning}
\end{abstract}

%% file: 01introduction.tex
\section{Introduction}
\label{label_introduction}

Federated learning (FL) is a decentralized approach to machine learning (ML) in which multiple devices or servers collaboratively train a shared global model while keeping the data on their local systems \cite{FedAvgM}. This method enhances privacy and data security by ensuring that raw data never leaves the local devices or servers \cite{truong_sun,huang_bert}. In our context, we refer to the devices as local ML models and to the central server as the global ML model. Each participant trains the model on their local data and shares only the model updates, such as gradients or model weights, with the global model. The global model then aggregates these updates to improve the overall model \cite{ek_portet_lalanda,qi_chiaro_guzzo}. Key features of FL include privacy preservation, decentralization, communication efficiency, federated averaging, and scalability \cite{guendouzi_ouchani}.

However, due to unbalanced class labels, each individual model may overfit or underfit specific labels when trained using a fixed number of epochs \cite{zhang_li,dust_murcia,wang_xu_wang}. Our primary objective is to improve the generalizability of the local models \cite{mora_bujari,sun_niu_wei}, thereby increasing classification accuracy through the use of early stopping, which enhances the aggregation step. By computing the discrepancy of the embeddings between the new class labels of the local and global models after each epoch, the number of epochs for specific local models with high generalization can be reduced. Consequently, local models can focus on fine-tuning challenging class labels, which decreases training time and increases model generalization.

\begin{figure}[!b]
    \centering
    \includegraphics[width=1.0\linewidth]{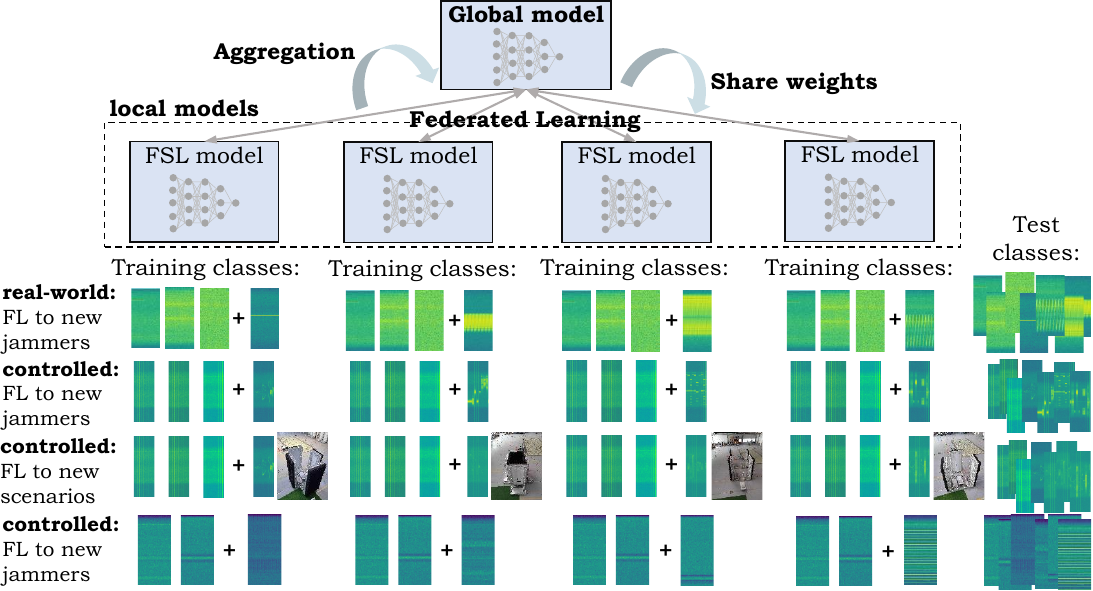}
    \caption{Overview of our FL method with aggregation and weight sharing steps between global and local models, aimed at adapting to new GNSS interference classes and scenarios in both real-world and controlled environments.}
    \label{figure_introduction_figure}
\end{figure}

Applications of FL include healthcare \cite{joshi_pal,rahman_hossain}, finance \cite{liu_wang}, and mobile devices \cite{li_liu_zhou,zou_feng,xiang_bi_chen}. This paper addresses an additional application: classifying interferences in global navigation satellite system (GNSS) signals \cite{brieger_ion_gnss,raichur_heublein,heublein_raichur_ion,heublein_feigl_crpa,heublein_feigl_posnav,manjunath_heublein}. Interference signals significantly disrupt GNSS signals, causing inaccuracies in localization and leading to errors in critical applications \cite{merwe_franco}. For instance, jammers on highways can launch jamming attacks that result in accidents involving self-driving cars \cite{zangvil_2019}. Our objective is to establish sensor stations along highways to record GNSS signals and detect and classify interferences. The local models at each station fine-tune on new interference classes and aggregate their model weights through a global model, enhancing the re-identification of new interference classes at all sensor stations \cite{ott_heublein_icl,raichur_ion_gnss}. Figure~\ref{figure_introduction_figure} provides an overview of our FL method: Local models employ few-shot learning (FSL), specifically a prototypical network \cite{snell_swersky_zemel}, to adapt to new classes or environmental scenarios and aggregate their weights \cite{wang_gong}.

The primary objective of this work is to propose an FL approach capable of adapting to novel interference classes to orchestrate sensor stations. Our contributions are outlined as follows: (1) Based on FedAvgM~\cite{FedAvgM}, we introduce our FL method, which utilizes FSL \cite{snell_swersky_zemel} and aggregation to adapt to new interferences. (2) We propose an early stopping approach to balance underrepresented datasets for greater model generalization. (3) We compute the discrepancy between local and global model embeddings, specifically based on the maximum mean discrepancy (MMD) \cite{borgwardt_gretton,long_zhu_mmd}. (4) Our evaluations, conducted on three GNSS datasets \cite{ott_heublein_icl,ott_heublein_dataset}, demonstrate that our FL method yields enhanced feature representation, resulting in improved classification accuracy. Our proposed MMD-based early stopping method is primarily designed for GNSS interference monitoring applications but can be generalized to a wide range of other use cases.

%The remainder of this paper is organized as follows. Section~\ref{label_related_work} provides an overview of the existing literature on FL and GNSS interference classification. Our proposed FL method is detailed in Section~\ref{label_methodology}. In Section~\ref{label_experiments}, we introduce three datasets of interference snapshots. Section~\ref{label_evaluation} summarizes the evaluation results, followed by the concluding remarks in Section~\ref{label_conclusion}.

%% file: 02related_work.tex
\section{Related Work}
\label{label_related_work}

In this section, we summarize state-of-the-art FL methods (see Section~\ref{label_rw_federated_learning}) and present related methods for GNSS interference classification (see Section~\ref{label_rw_gnss}).

\subsection{Federated Learning (FL)}
\label{label_rw_federated_learning}

Our MMD-based FL approach aligns with the IEEE NOMS community's focus on federated and model-driven management, contributing to advancements in ML methods for managing complex and distributed networks and services. FL enables ML models to adapt to local environments or specific purposes by preprocessing information on local, distributed nodes. Sattler et al.~\cite{sattler2019robust} demonstrated that compression methods, when orchestrated, provide excellent accuracy on independent and identically distributed (i.i.d.) data. However, as performance degrades with differing data distributions, Hsu et al.~\cite{FedAvgM} proposed federated averaging with momentum (FedAvgM) to synthesize datasets with a continuous range of similarity and introduced a mitigation strategy, via server momentum. Yin et al.~\cite{FedMedian} developed distributed optimization algorithms that are robust against Byzantine failures (arbitrary and potentially adversarial behavior) by utilizing gradient descent methods based on median and trimmed mean operations. Wang et al.~\cite{FedtrimmedAvg} proposed FedTrimmedAvg, which utilizes a variant of the Shapley value (SV) that preserves the desirable properties of the canonical SV, while being calculable without extra communication costs. SV also captures the effect of participation order on data value. Decentralized FedAvg with momentum (DFedAvgM)~\cite{Fedavg} connects clients by an undirected graph, with all clients performing stochastic gradient descent with momentum and communicating only with their neighbors. FedProx, introduced by Li et al.~\cite{FedProx}, addresses heterogeneity in federated networks to generalize and re-parameterize FedAvg. Given that FedAvg is often difficult to tune and exhibits unfavorable convergence behavior, FedOpt~\cite{FedOpt} integrates federated versions of adaptive optimizers, including AdaGrad~\cite{duchi_hazan}, Adam~\cite{kingma_ba}, and YOGI~\cite{zaheer_reddi}, to enhance client heterogeneity and communication efficiency. Our focus, however, is not on the federated optimizer, but on improving the convergence behavior of FedAvg through optimal class balance. FaultToleranceFedAvg~\cite{faulttolerant}, based on Zeno, generalizes previous results by assuming a majority of non-faulty nodes, allowing potentially defective workers to be identified through a ranking-based preference mechanism. Shi et al.~\cite{Shi} presented an FL framework for signal modulation detection that ensures privacy and security, while Salama et al.~\cite{Abdelaziz} proposed a robust wireless communication network for FL. The adaptive collaborative FL method proposed by Xiang et al.~\cite{xiang_bi_chen} enhances convergence speed and improves model reliability by addressing communication-related parameter loss. While their approach of dynamic epoch adjustment reduces the number of communication rounds, it necessitates the computation of additional, unspecified parameters. Additional early stopping methods include those proposed by \cite{niu_dong_qin,li_wang_zhang,jiang_ying,zeng_yang_chen,mcmahan_moore}. In summary, FL still lacks a well-defined training process for unbalanced datasets and has not yet addressed applications for orchestrating GNSS interference classification models.

\subsection{GNSS Interference Classification}
\label{label_rw_gnss}

\begin{figure*}[!t]
    \centering
    \includegraphics[width=0.93\linewidth]{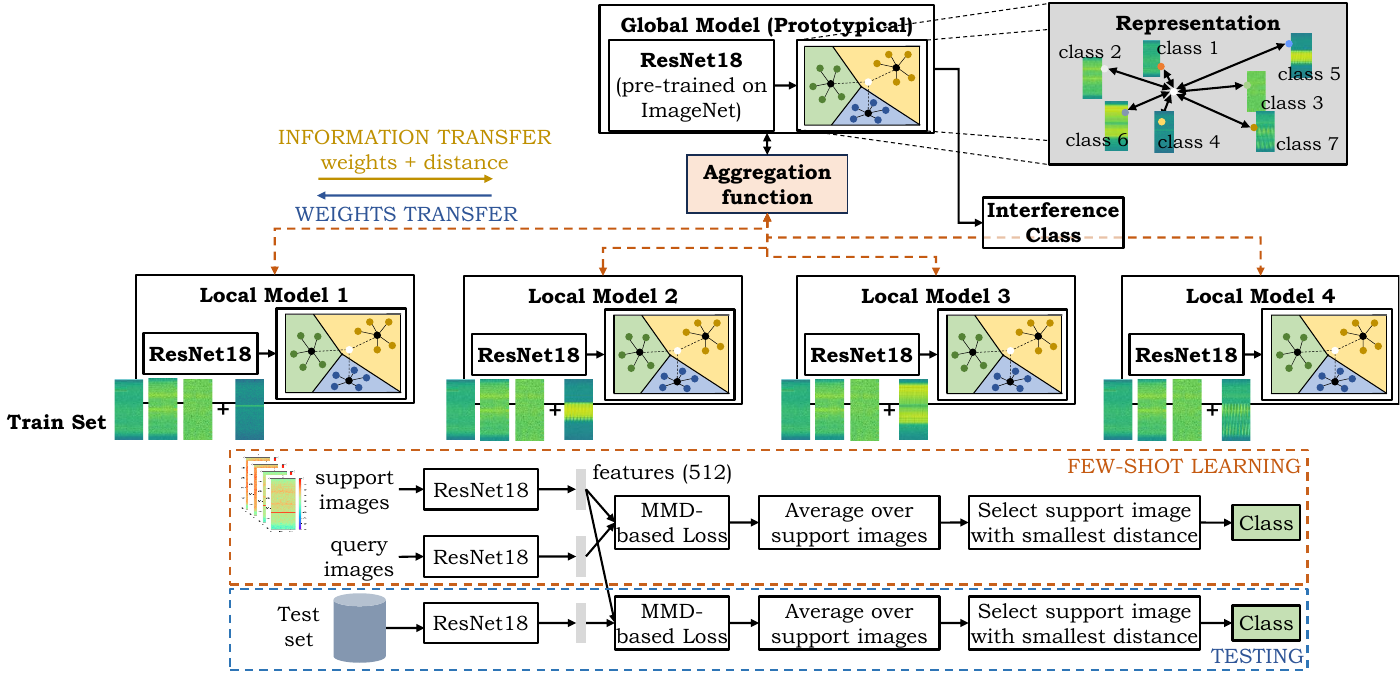}
    \caption{\textbf{Overview of our FL method.} The global model is based on a prototypical FSL method utilizing ResNet18 to learn an optimal feature representation. The global model aggregates and transfers weights from several local models. Each local model fine-tunes on a new interference class. The bottom pipeline shows the FSL process with MMD-based early stopping.}
    \label{figure_method_overview}
\end{figure*}

Recently, ML methods have been employed to analyze and process GNSS interference signals. Ferre et al.~\cite{s19224841} introduced ML techniques, such as support vector machines and convolutional neural networks, to classify jammer types in GNSS signals. Ding et al.~\cite{Ding} exploited ML models in a single static (line-of-sight) propagation environment. Gross et al.~\cite{Gross} applied a maximum likelihood method to determine whether a synthetic signal is affected by multipath or jamming. However, their method is not capable of classifying real-world multipath effects and jammer types. Jdidi et al.~\cite{jdidi_brieger} proposed an unsupervised method to adapt to diverse, environment-specific factors, such as multipath effects, dynamics, and variations in signal strength. However, their pipeline requires human intervention to monitor the process and address uncertain classifications. While Taori et al.~\cite{taori_dave} evaluated the distributional shift from synthetic to real data, our evaluation focuses on assessing model robustness in adapting to different classes and scenarios within real-world and controlled environments. In recent developments, FSL~\cite{wang_yao,narwariya_malhotra,luo_si,tang_liu_long} has been used in the GNSS context to integrate new classes into a support set. Ott et al.~\cite{ott_heublein_icl,heublein_feigl_posnav} proposed an uncertainty-based quadruplet loss for prototypical networks \cite{snell_swersky_zemel}, aiming for a more continuous representation between positive and negative interference pairs. They utilized a dataset recorded along a highway, resembling a snapshot-based real-world dataset featuring similar interference classes. Raichur et al.~\cite{raichur_heublein} used the same dataset to adapt to novel interference classes through continual learning. Brieger et al.~\cite{brieger_ion_gnss} integrated both the spatial and temporal relationships between samples by using a joint loss function and a late fusion technique. Their dataset was acquired in a controlled indoor small-scale environment. As ResNet18~\cite{he_zhang} is widely used for feature extraction and proved robust for interference classification due to its efficient architecture and performance, we also employ ResNet18 for feature extraction. Raichur et al.~\cite{raichur_ion_gnss} introduced a crowdsourcing method utilizing smartphone-based features to localize the source of detected interference.

%% file: 03method.tex
\section{Methodology}
\label{label_methodology}

\textbf{Overview.} Figure~\ref{figure_method_overview} provides a comprehensive overview of our FL method for GNSS interference classification. Initially, the global model ResNet18~\cite{he_zhang} is pre-trained on ImageNet~\cite{deng_dong_socher} and fine-tuned on a subset of classes from the GNSS dataset using the cross-entropy (CE) loss to predict interference classes. The initial weights are provided to the local models. The goal is to learn a continuous feature representation for all classes. The global model and its weights are then distributed to all local models. Each local model is fine-tuned on a novel interference class alongside previously seen classes. However, there is a discrepancy between the data distributions of the previous and novel interference classes. The models are adapted using a prototypical FSL network~\cite{snell_swersky_zemel}, depicted in the bottom pipeline of Figure~\ref{figure_method_overview}. After class adaptation, the weights and embedding distances are sent to the global model for an aggregation step based on FedAvgM~\cite{FedAvgM}. Following aggregation, the updated weights and epoch parameters are sent back to the local models. Subsequently, we provide a detailed description of the FSL approach. Our main contribution is the implementation of early stopping with dynamic epoch selection based on class discrepancy.

\textbf{Notation.} Let $\mathbf{X} \in \mathbb{R}^{h \times w}$ with entries $x_{i,j} \in [0, 255]$ with height $h$ and width $w$, represents an image from the image training set. The images are obtained by calculating the magnitude spectrogram of snapshots of IQ-samples. The image training set is a subset of the array $\mathcal{X} = \{\mathbf{X}_1,\ldots,\mathbf{X}_{n_X}\} \in \mathbb{R}^{n_X \times h \times w}$, where $n_X$ is the number of images in the training set. The goal is to predict an unknown class label $y \in \Omega$. We define the classification task as a multi-class task with one (or more) '\textit{no interference}' classes with varying background intensity and several '\textit{interference}' classes, e.g., '\textit{chirp}' \cite{ott_heublein_icl,brieger_ion_gnss}. The goal is to learn representative feature embeddings $f(\mathbf{X}) \in \mathbb{R}^{q \times t}$ to map the image input into a feature space $\mathbb{R}^{q \times t}$, where $f$ are the output of a specified model layer and $q \times t$ is the dimension of the layer output.

\textbf{Few-Shot Learning.} The primary objective is to learn a representation that minimizes intra-class distances while maximizing inter-class distances. The prototypical network~\cite{snell_swersky_zemel} employs both a support and a query set. During the adaptation phase, the distance between features, typically measured using the Euclidean metric, is computed from the 512-dimensional outputs of the final layer for both the support and query image. At inference, the class assigned to the query image is determined by the smallest distance to the average of the support images. These feature embeddings are utilized in our early stopping method, which is detailed in the subsequent sections.

\begin{table}[!t]
\caption{Overview of our proposed FL algorithm.}
\label{tab:FedAvg}
\begin{tabular}{>{\raggedright\arraybackslash}p{0.95\linewidth}p{0.95\linewidth}}
\toprule
\textbf{Federated learning with MMD-based early stopping.} \\
\midrule
\textbf{Input:} Number of local models $M$; local models $h_i(\omega)$, $i \in M$;
global model $h_g(\omega)$; initial weights $\omega_0$; dataset $D = \{D_1, \ldots, D_M\}$;
number of rounds $R$; number of epochs $E_{\text{min},i,r}$ and $E_{\text{max},i,r}$;
learning rate $\eta$ \\
Initialize global model $h_g(\omega)$ with $\omega_0$ \\
Transfer global model's weights $h_g$ to all local models $h_i, \forall i \in M$ \\
Train local models $h_i(\omega_0)$ with train sets $D_i, \forall i \in M$ \\
\textbf{for $r = 1, \ldots, R$} \\
\quad \textbf{for $i = 1, \ldots, M$} \\
\qquad \textbf{if $r <= 2$} \\
\qquad \quad $h_i(\omega_{r}) \leftarrow$ solution of training local model $h_i(\omega_{r-1})$ with \\
\qquad \qquad data $D_i$ for epochs $E_{\text{max},i,r}$ with learning rate $\eta$ \\
\qquad \textbf{else} \\
\qquad \quad $h_i(\omega_{r}) \leftarrow$ solution of training local model $h_i(\omega_{r-1})$ with \\
\qquad \qquad data $D_i$ for epochs $E_{i,r}$ with learning rate $\eta$ \\
\qquad \textbf{endIf} \\
\qquad Upload local model weights $\omega_r$ of $h_i$ and embedding loss to \\
\qquad \quad global model $h_g(\omega_r)$ \\
\qquad Compute MMD between $f(h_g(\omega_0))$ and $f(h_i(\omega_{r-1}))$ \\
\qquad \quad for all $i \in M$ with Eq.~(\ref{equ3}) \\
\qquad Calculate epoch $E_{i,r} \in [E_{\text{min},i,r}, E_{\text{max},i,r}]$ with Eq.~(\ref{equ1}) and \\
\qquad \quad Eq.~(\ref{equ2}) \\
\quad \textbf{endFor} \\
\quad Aggregation: average local model's weights $h_i(\omega_{r-1})$ \\
\quad Update global model weights $h_g(\omega_r)$ \\
\quad Share weights: send and update global model weights $h_g(\omega_r)$ to \\
\qquad local models $h_i(\omega_r), \forall i \in M$ \\
\textbf{endFor} \\
\textbf{Output:} $h_g(\omega_0)$, $h_i(\omega_r), \forall i \in M$ \\
\bottomrule
\end{tabular}
\end{table}

\textbf{Early Stopping.} As further explained in Section~\ref{label_experiments}, certain classes exhibit similar appearances, resulting in similar feature representations, whereas other classes display significant discrepancies; for example, the classes '\textit{chirp}' and '\textit{single tone}'. Additionally, the dataset is imbalanced, containing over 100,000 negative samples compared to only a few hundred positive samples. To address this, we implement an early stopping mechanism based on the discrepancy between feature embeddings. We define the number of rounds $R$ as the number of complete communication cycles between the local and global models. The number of epochs $E \in [E_{\textbf{min}}, E_{\textbf{max}}]$ represents the full training iterations over the local dataset for a local model. We denote $E_{i,r}$ as the number of epochs for local model $i$ at round $r$. After the aggregation step, if the feature discrepancy is low, we reduce the number of epochs per round; conversely, if the feature discrepancy is high, we train up to $E_{\textbf{max}} = 5$ epochs per round. Let $M$ represent the number of local models $h_i(\omega)$, where $i \in \{1, 2, \ldots, M\}$ and $\omega$ are the model weights. The global model is denoted as $h_g(\omega)$. Algorithm~\ref{tab:FedAvg} outlines the steps of our FL method. For each round and each model, the local models are first initialized and trained on their respective datasets $D_i$ for $E_{i,r}$ epochs. Subsequently, the model weights $\omega_r$ of $h_i(\omega_r)$ and the embedding distance $d$ are transmitted to the global model, where the discrepancy between the embeddings $f(h_g(\omega_{0}))$ and $f(h_i(\omega_r))$ is computed. Based on this discrepancy, we determine the maximum number of epochs $E_{i,r}$ for each model for the next round $r$ with
\begin{equation}
\label{equ1}
    \resizebox{1.0\linewidth}{!}{$
    \displaystyle
    E_{\text{max},i,r} = 
    \begin{cases}
    E_{\text{max},i,r} & \text{for } r \leq 2 \\
    \max\left(\left\lfloor \frac{\max\Big(d\big(f(h_g(\omega_{0})) - f(h_i(\omega_{r}))\big)\Big)}{\max\Big(d\big(f(h_g(\omega_{0})) - f(h_i(\omega_{r-1}))\big)\Big)} \cdot E_{\text{max},i,r} \right\rfloor, E_{\text{min},i,r}\right) & \text{for } r > 2 
    \end{cases}
    $}
\end{equation}
For each model, the number of epochs are calculated by
\begin{equation}
    \label{equ2}
    \resizebox{0.90\linewidth}{!}{$
    \displaystyle
    E_{i,r} = \left\lfloor E_{\text{max},i,r} \cdot \frac{d\big(f(h_g(\omega_{0})) - f(h_i(\omega_{r}))\big)}{\max\Big(d\big(f(h_g(\omega_{0})) - f(h_i(\omega_{r}))\big)\Big)} \right\rfloor.
    $}
\end{equation}
Finally, the model weights are aggregated \cite{FedAvgM}; the global model weights are updated, and the updated weights are distributed to the local models.

\textbf{Class Discrepancy.} A standard metric to compute the distance between feature embeddings is the mean squared error (MSE). However, MSE is highly sensitive to the scale and magnitude of the feature embeddings and is non-robust to outliers \cite{huang_peng}. In recent years, the maximum mean discrepancy (MMD)~\cite{borgwardt_gretton,long_zhu_mmd} loss has emerged as a standard metric for domain adaptation~\cite{ott_acmmm} and cross-modal retrieval~\cite{ott_access} applications. Hence, we compute the discrepancy between feature embeddings $d$ in Eq.~(\ref{equ2}) using
\begin{equation}
    \label{equ3}
    \resizebox{1.0\linewidth}{!}{$
    \displaystyle
    d\big(f(h_g(\omega_{0})), f(h_i(\omega_{r}))\big) = \bigg|\bigg| \frac{1}{n_{h_g}} \sum_{k=1}^{n_{h_g}} f(h_g^k(\omega_{0})) - \frac{1}{n_{h_i}} \sum_{k=1}^{n_{h_i}} f(h_i^k(\omega_{r})) \bigg|\bigg|^2,
    $}
\end{equation} 
where $h_g^k$ and $h_i^k$ are the $k$-th entries of $h_g$ and $h_i$, respectively, and $n_{h_g}$ and $n_{h_i}$ are the lengths of $h_g$ and $h_i$, respectively~\cite{borgwardt_gretton}. Higher loss values at local models result in a higher number of epochs being assigned, reflecting the increased difficulty.

\textbf{Computational Complexity.} The model weights for the ResNet18 model have a size of 45 MB. We train the $M$ local models over $R$ rounds, with each round consisting of 5 epochs. Given the bidirectional communication overhead of transferring weights, the total communication overhead for state-of-the-art methods amounts to $2 \times M \times R \times 5 \times 45 \, \text{MB}$. However, by dynamically adjusting the maximum number of epochs through early stopping, our communication overhead is reduced to $2 \times M \times R \times E_{max} \times 45 \, \text{MB}$.

%% file: 04experiments.tex
\section{Experiments}
\label{label_experiments}

\newcommand\y{0.079}
\newcommand\x{0.087}
\newcommand\z{0.14}
\begin{figure}[!t]
    \centering
	\begin{minipage}[t]{\y\linewidth}
        \centering
    	\includegraphics[trim=150 40 90 45, clip, width=1.0\linewidth]{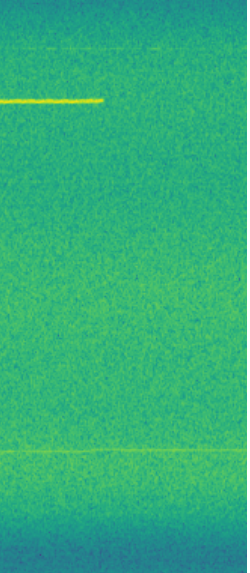}
    	\subcaption{None}
    	\label{figure_exemplary_samples0}
    \end{minipage}
    \hfill
	\begin{minipage}[t]{\y\linewidth}
        \centering
    	\includegraphics[trim=150 40 90 45, clip, width=1.0\linewidth]{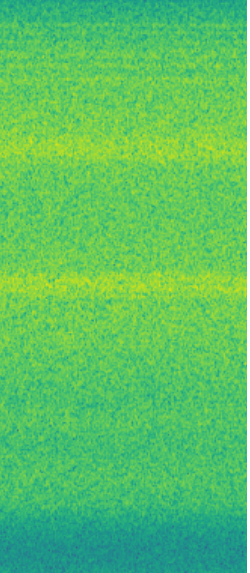}
    	\subcaption{None}
    	\label{figure_exemplary_samples1}
    \end{minipage}
    \hfill
	\begin{minipage}[t]{\y\linewidth}
        \centering
    	\includegraphics[trim=150 40 90 45, clip, width=1.0\linewidth]{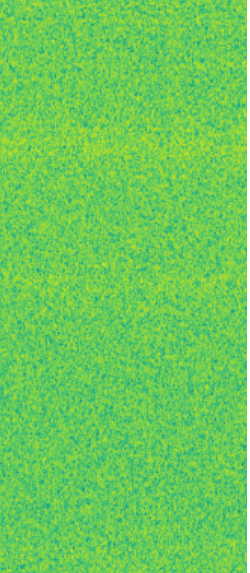}
    	\subcaption{None}
    	\label{figure_exemplary_samples2}
    \end{minipage}
    \hfill
	\begin{minipage}[t]{\y\linewidth}
        \centering
    	\includegraphics[trim=150 40 90 45, clip, width=1.0\linewidth]{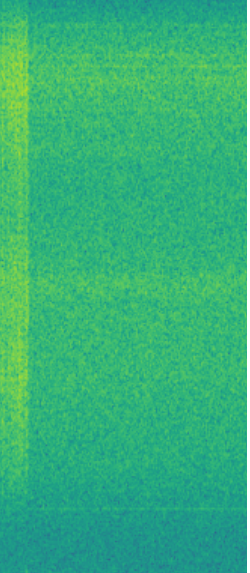}
    	\subcaption{Puls}
    	\label{figure_exemplary_samples3}
    \end{minipage}
    \hfill
	\begin{minipage}[t]{\y\linewidth}
        \centering
    	\includegraphics[trim=150 40 90 45, clip, width=1.0\linewidth]{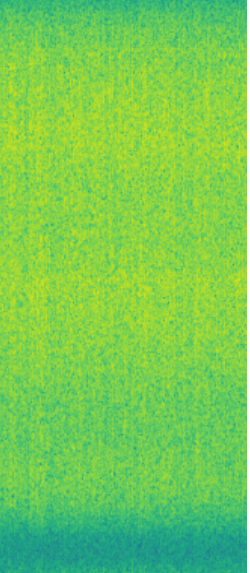}
    	\subcaption{Puls}
    	\label{figure_exemplary_samples4}
    \end{minipage}
    \hfill
	\begin{minipage}[t]{\y\linewidth}
        \centering
    	\includegraphics[trim=150 40 90 45, clip, width=1.0\linewidth]{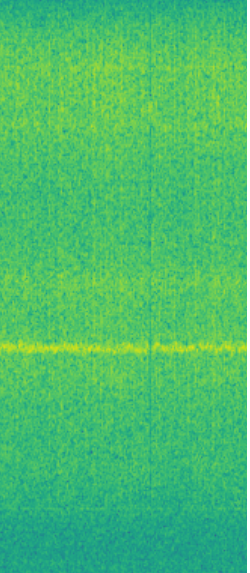}
    	\subcaption{Tone}
    	\label{figure_exemplary_samples5}
    \end{minipage}
    \hfill
	\begin{minipage}[t]{\y\linewidth}
        \centering
    	\includegraphics[trim=150 40 90 45, clip, width=1.0\linewidth]{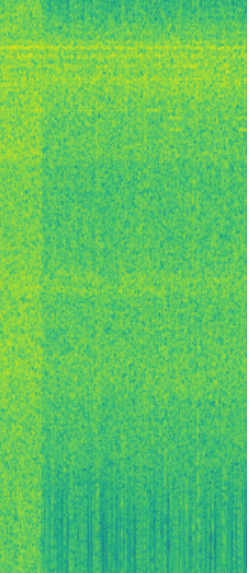}
    	\subcaption{Noise}
    	\label{figure_exemplary_samples6}
    \end{minipage}
    \hfill
	\begin{minipage}[t]{\y\linewidth}
        \centering
    	\includegraphics[trim=150 40 90 45, clip, width=1.0\linewidth]{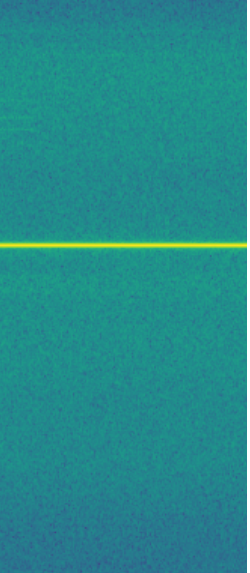}
    	\subcaption{Tone}
    	\label{figure_exemplary_samples7}
    \end{minipage}
    \hfill
	\begin{minipage}[t]{\y\linewidth}
        \centering
    	\includegraphics[trim=150 40 90 45, clip, width=1.0\linewidth]{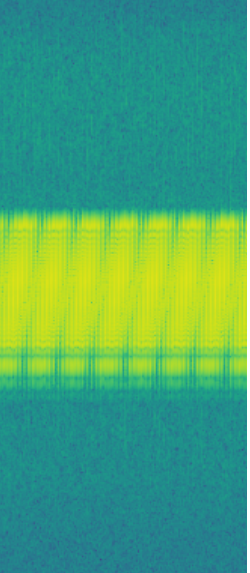}
    	\subcaption{Chirp}
    	\label{figure_exemplary_samples8}
    \end{minipage}
    \hfill
	\begin{minipage}[t]{\y\linewidth}
        \centering
    	\includegraphics[trim=150 40 90 45, clip, width=1.0\linewidth]{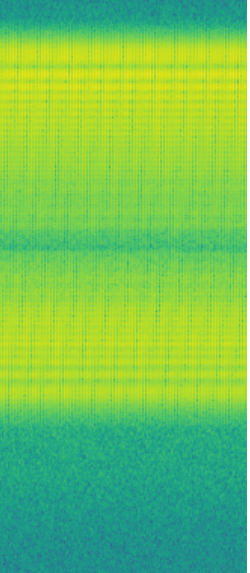}
    	\subcaption{2 chirps}
    	\label{figure_exemplary_samples9}
    \end{minipage}
    \hfill
	\begin{minipage}[t]{\y\linewidth}
        \centering
    	\includegraphics[trim=150 40 90 45, clip, width=1.0\linewidth]{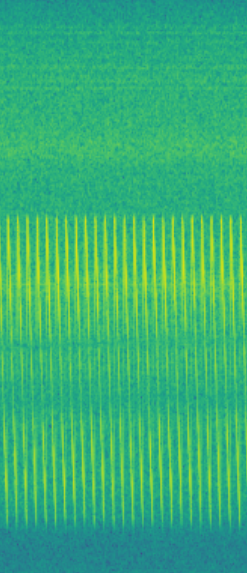}
    	\subcaption{Chirp}
    	\label{figure_exemplary_samples10}
    \end{minipage}
	\begin{minipage}[t]{\x\linewidth}
        \centering
    	\includegraphics[trim=170 55 90 45, clip, width=1.0\linewidth]{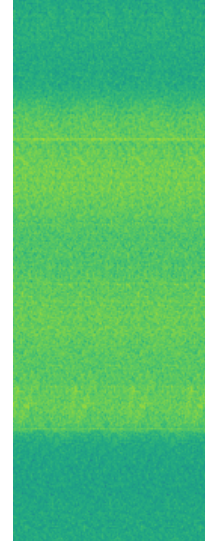}
    	\subcaption{None}
    	\label{figure_controlled_short0}
    \end{minipage}
    \hfill
	\begin{minipage}[t]{\x\linewidth}
        \centering
    	\includegraphics[trim=170 55 90 45, clip, width=1.0\linewidth]{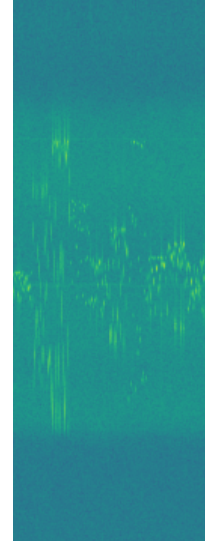}
    	\subcaption{Chirp, sc. 1}
    	\label{figure_controlled_short1}
    \end{minipage}
    \hfill
	\begin{minipage}[t]{\x\linewidth}
        \centering
    	\includegraphics[trim=170 55 90 45, clip, width=1.0\linewidth]{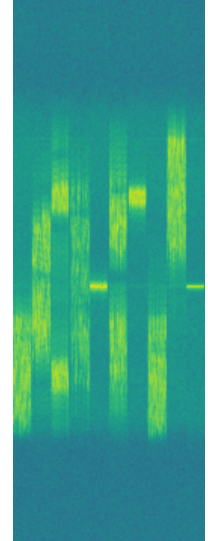}
    	\subcaption{FH}
    	\label{figure_controlled_short2}
    \end{minipage}
    \hfill
	\begin{minipage}[t]{\x\linewidth}
        \centering
    	\includegraphics[trim=170 55 90 45, clip, width=1.0\linewidth]{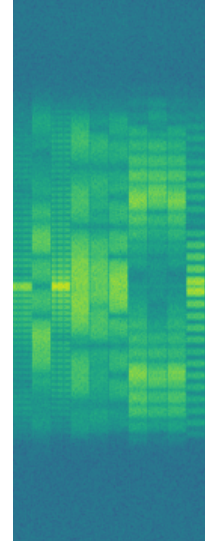}
    	\subcaption{Modul.}
    	\label{figure_controlled_short3}
    \end{minipage}
    \hfill
	\begin{minipage}[t]{\x\linewidth}
        \centering
    	\includegraphics[trim=170 55 90 45, clip, width=1.0\linewidth]{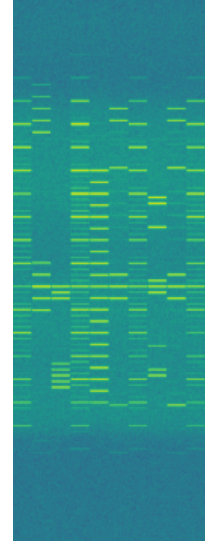}
    	\subcaption{Multitone}
    	\label{figure_controlled_short4}
    \end{minipage}
    \hfill
	\begin{minipage}[t]{\x\linewidth}
        \centering
    	\includegraphics[trim=170 55 90 45, clip, width=1.0\linewidth]{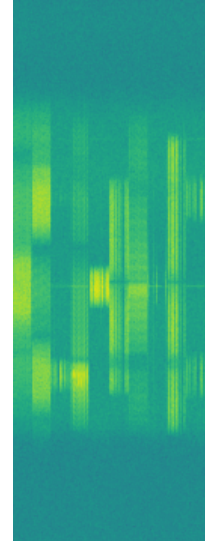}
    	\subcaption{Puls}
    	\label{figure_controlled_short5}
    \end{minipage}
    \hfill
	\begin{minipage}[t]{\x\linewidth}
        \centering
    	\includegraphics[trim=170 55 90 45, clip, width=1.0\linewidth]{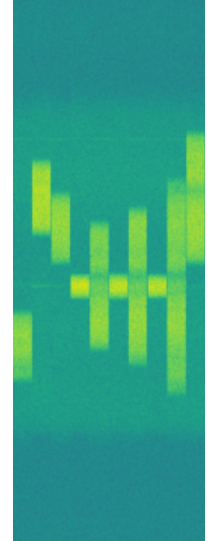}
    	\subcaption{Noise}
    	\label{figure_controlled_short6}
    \end{minipage}
    \hfill
	\begin{minipage}[t]{\x\linewidth}
        \centering
    	\includegraphics[trim=170 55 90 45, clip, width=1.0\linewidth]{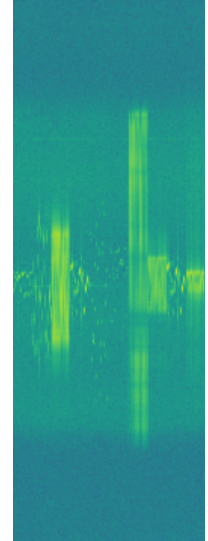}
    	\subcaption{Chirp, sc. 3}
    	\label{figure_controlled_short7}
    \end{minipage}
    \hfill
	\begin{minipage}[t]{\x\linewidth}
        \centering
    	\includegraphics[trim=170 55 90 45, clip, width=1.0\linewidth]{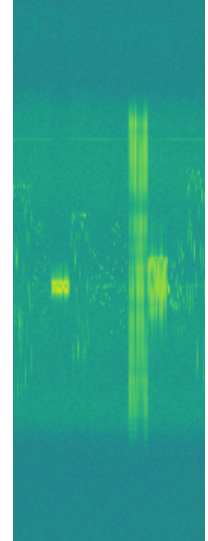}
    	\subcaption{Chirp, sc. 6}
    	\label{figure_controlled_short8}
    \end{minipage}
    \hfill
	\begin{minipage}[t]{\x\linewidth}
        \centering
    	\includegraphics[trim=170 55 90 45, clip, width=1.0\linewidth]{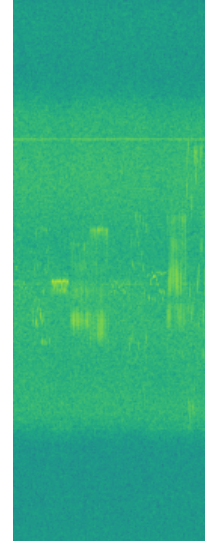}
    	\subcaption{Chirp, sc. 8}
    	\label{figure_controlled_short9}
    \end{minipage}
	\begin{minipage}[t]{\z\linewidth}
        \centering
    	\includegraphics[trim=150 30 36 38, clip, width=1.0\linewidth]{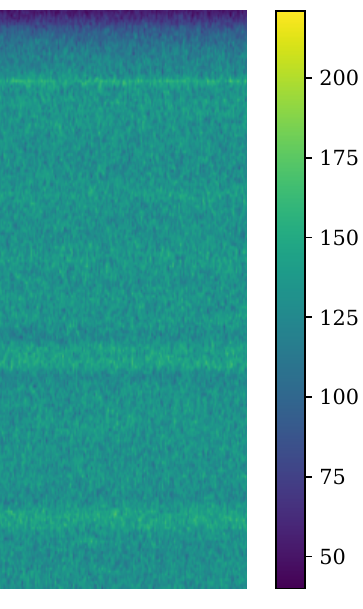}
    	\subcaption{None}
    	\label{figure_controlled_GNSS0}
    \end{minipage}
    \hfill
	\begin{minipage}[t]{\z\linewidth}
        \centering
    	\includegraphics[trim=150 30 36 38, clip, width=1.0\linewidth]{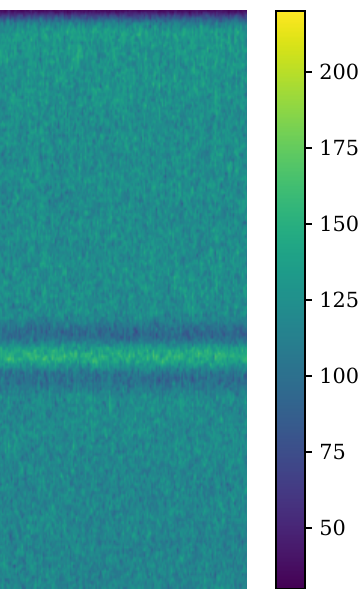}
    	\subcaption{1}
    	\label{figure_controlled_GNSS1}
    \end{minipage}
    \hfill
	\begin{minipage}[t]{\z\linewidth}
        \centering
    	\includegraphics[trim=150 30 36 38, clip, width=1.0\linewidth]{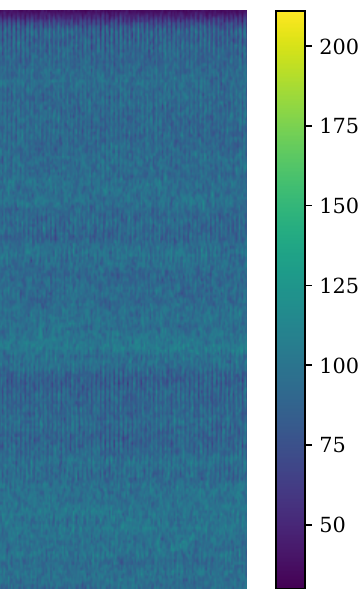}
    	\subcaption{2}
    	\label{figure_controlled_GNSS2}
    \end{minipage}
    \hfill
	\begin{minipage}[t]{\z\linewidth}
        \centering
    	\includegraphics[trim=150 30 36 38, clip, width=1.0\linewidth]{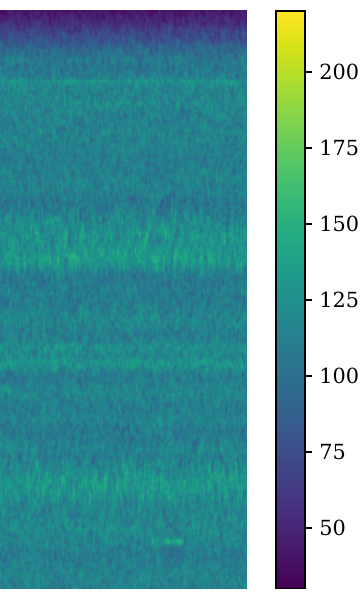}
    	\subcaption{3}
    	\label{figure_controlled_GNSS3}
    \end{minipage}
    \hfill
	\begin{minipage}[t]{\z\linewidth}
        \centering
    	\includegraphics[trim=150 30 36 38, clip, width=1.0\linewidth]{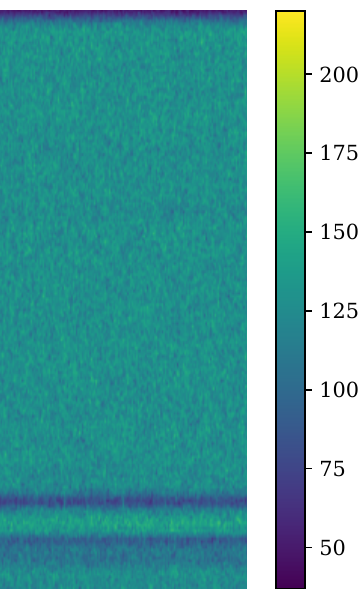}
    	\subcaption{4}
    	\label{figure_controlled_GNSS4}
    \end{minipage}
    \hfill
	\begin{minipage}[t]{\z\linewidth}
        \centering
    	\includegraphics[trim=150 30 36 38, clip, width=1.0\linewidth]{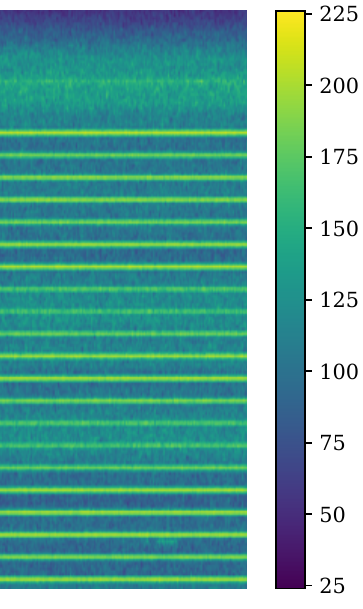}
    	\subcaption{5}
    	\label{figure_controlled_GNSS5}
    \end{minipage}
    \caption{Exemplary spectrogram samples recorded in a real-world highway environment (top), controlled large-scale environment (middle), and controlled small-scale environment (bottom). The x-axis shows the time in $\text{ms}$. The y-axis shows the frequency in $\text{MHz}$.}
    \label{figure_exemplary_samples}
\end{figure}

In this section, we present three datasets for evaluating FL in GNSS interference classification, with the primary objective of adapting to new interference classes and new environments. We introduce one real-world highway dataset and two datasets recorded in different indoor environments with controlled multipath effects. The datasets are available at \cite{ott_heublein_dataset}. Figure~\ref{figure_exemplary_samples} shows exemplary spectrogram samples with intensity values ranging from [0, -90] on a logarithmic scale for all datasets, highlighting various interference types. Table~\ref{table_dataset_overview} summarizes the class labels on which the global model is initially trained, the adaptation set of the local models, and the test set, including environmental labels.

\textbf{Real-world Highway Dataset.} For the first dataset~\cite{ott_heublein_icl}, a sensor station was placed on a bridge along a highway to capture short, wideband snapshots in both the E1 and E6 GNSS bands. The setup records 20\,ms raw IQ snapshots triggered by energy, with a sample rate of 62.5\,MHz, an analog bandwidth of 50\,MHz, and an 8\,bit width. The spectrogram images have dimensions of $512 \times 243$. Manual labeling of these snapshots resulted in 11 classes: Classes 0 to 2 represent samples with no interferences (Figure~\ref{figure_exemplary_samples0} to \ref{figure_exemplary_samples2}), distinguished by variations in background intensity, while classes 3 to 10 contain different interferences (Figure~\ref{figure_exemplary_samples3} to \ref{figure_exemplary_samples10}). The dataset is highly imbalanced, presenting a challenge in adapting to positive class labels with only a limited number of samples available. Moreover, a second dataset was collected at a different highway location using the same sensor station~\cite{heublein_raichur_ion}. This dataset comprises nine classes and includes a total of 16,480 samples. The evaluation focuses on the adaptation to new interference classes between the two highway locations.

\textbf{Controlled Large-scale Dataset.} A second GNSS dataset was recorded using a low-frequency antenna in a controlled large-scale indoor environment~\cite{heublein_feigl_crpa}, capturing six interference classes (Figure~\ref{figure_controlled_short1} to \ref{figure_controlled_short6}) with a total of 42,592 snapshot samples. The spectrogram images have dimensions of $1,024 \times 34$. We evaluate the adaptation to three specific class labels (see Table~\ref{table_dataset_overview}, bottom). Figure~\ref{figure_controlled_short1}, \ref{figure_controlled_short7}, \ref{figure_controlled_short8}, and \ref{figure_controlled_short9} demonstrate a '\textit{Chirp}' in four different scenarios, each with varying amounts of multipath effects. For the adaptation to new scenarios, we group two different challenging sets: one with fewer multipath effects for model 1 and another with higher multipath effects for model 2.

\begin{table}[!t]
\begin{center}
    \caption{Overview of class and environmental labels for the real-world highway environment (top), the controlled small-scale environment (middle), and controlled large-scale environment (bottom; class labels 0, 1, 2, and 3) for the training, adaptation (in \textbf{bold}), and test set.}
    \label{table_dataset_overview}
    \begin{tabular}{ p{0.5cm} | p{0.5cm} | p{0.5cm} | p{0.5cm} }
    \multicolumn{1}{c|}{\textbf{Model}} & \multicolumn{1}{c|}{\textbf{Training}} & \multicolumn{1}{c|}{\textbf{Adaptation}} & \multicolumn{1}{c}{\textbf{Test}} \\ \hline
    \multicolumn{1}{l|}{Global} & \multicolumn{1}{c|}{0, 1, 2, 4, 5, 7, 10} & \multicolumn{1}{c|}{-} & \multicolumn{1}{r}{all} \\
    \multicolumn{1}{l|}{Local 1} & \multicolumn{1}{c|}{initialized} & \multicolumn{1}{c|}{0, 1, 2, \textbf{3}, 4, 5, 7, 10} & \multicolumn{1}{r}{all} \\
    \multicolumn{1}{l|}{Local 2} & \multicolumn{1}{c|}{initialized} & \multicolumn{1}{c|}{0, 1, 2, 4, 5, \textbf{6}, 7, 10} & \multicolumn{1}{r}{all} \\
    \multicolumn{1}{l|}{Local 3} & \multicolumn{1}{c|}{initialized} & \multicolumn{1}{c|}{0, 1, 2, 4, 5, 7, \textbf{8}, 10} & \multicolumn{1}{r}{all} \\
    \multicolumn{1}{l|}{Local 4} & \multicolumn{1}{c|}{initialized} & \multicolumn{1}{c|}{0, 1, 2, 4, 5, 7, \textbf{9}, 10} & \multicolumn{1}{r}{all} \\ \hline
    \multicolumn{1}{c|}{\textbf{Model}} & \multicolumn{1}{c|}{\textbf{Training}} & \multicolumn{1}{c|}{\textbf{Adaptation}} & \multicolumn{1}{c}{\textbf{Test}} \\ \hline
    \multicolumn{1}{l|}{Global} & \multicolumn{1}{c|}{0, 1, 2} & \multicolumn{1}{c|}{-} & \multicolumn{1}{r}{all} \\
    \multicolumn{1}{l|}{Local 1} & \multicolumn{1}{c|}{initialized} & \multicolumn{1}{c|}{0, 1, 2, \textbf{3}} & \multicolumn{1}{r}{all} \\
    \multicolumn{1}{l|}{Local 2} & \multicolumn{1}{c|}{initialized} & \multicolumn{1}{c|}{0, 1, 2, \textbf{4}} & \multicolumn{1}{r}{all} \\
    \multicolumn{1}{l|}{Local 3} & \multicolumn{1}{c|}{initialized} & \multicolumn{1}{c|}{0, 1, 2, \textbf{5}} & \multicolumn{1}{r}{all} \\ \hline
    \multicolumn{1}{c|}{} & \multicolumn{1}{c|}{\textbf{Training}} & \multicolumn{1}{c|}{\textbf{Adaptation}} & \multicolumn{1}{c}{} \\
    \multicolumn{1}{c|}{\textbf{Model}} & \multicolumn{1}{c|}{\textbf{environment}} & \multicolumn{1}{c|}{\textbf{environments}} & \multicolumn{1}{c}{\textbf{Test}} \\ \hline
    \multicolumn{1}{l|}{Global} & \multicolumn{1}{c|}{1} & \multicolumn{1}{c|}{-} & \multicolumn{1}{r}{all} \\
    \multicolumn{1}{l|}{Local 1} & \multicolumn{1}{c|}{initialized} & \multicolumn{1}{c|}{\textbf{2, 3, 4, 5, 11}} & \multicolumn{1}{r}{all} \\
    \multicolumn{1}{l|}{Local 2} & \multicolumn{1}{c|}{initialized} & \multicolumn{1}{c|}{\textbf{6, 7, 8, 9, 10}} & \multicolumn{1}{r}{all} \\
    \end{tabular}
\end{center}
\end{table}

\textbf{Controlled Small-scale Dataset.} The third dataset~\cite{brieger_ion_gnss} was recorded using the same high-frequency antenna as in the first dataset but in a controlled small-scale environment. This dataset contains five interference classes (Figure~\ref{figure_controlled_GNSS1} to \ref{figure_controlled_GNSS5}). The spectrogram images have dimensions of $512 \times 243$. For each local model, we adapt to three different classes (see Table~\ref{table_dataset_overview}, middle).

%% file: 05evaluation.tex
\section{Evaluation}
\label{label_evaluation}

\setlength{\intextsep}{6pt}
\setlength{\columnsep}{12pt}
\begin{wrapfigure}{R}{4.6cm}
    \begin{minipage}[b]{1.0\linewidth}
        \includegraphics[trim=10 10 10 10, clip, width=1.0\linewidth]{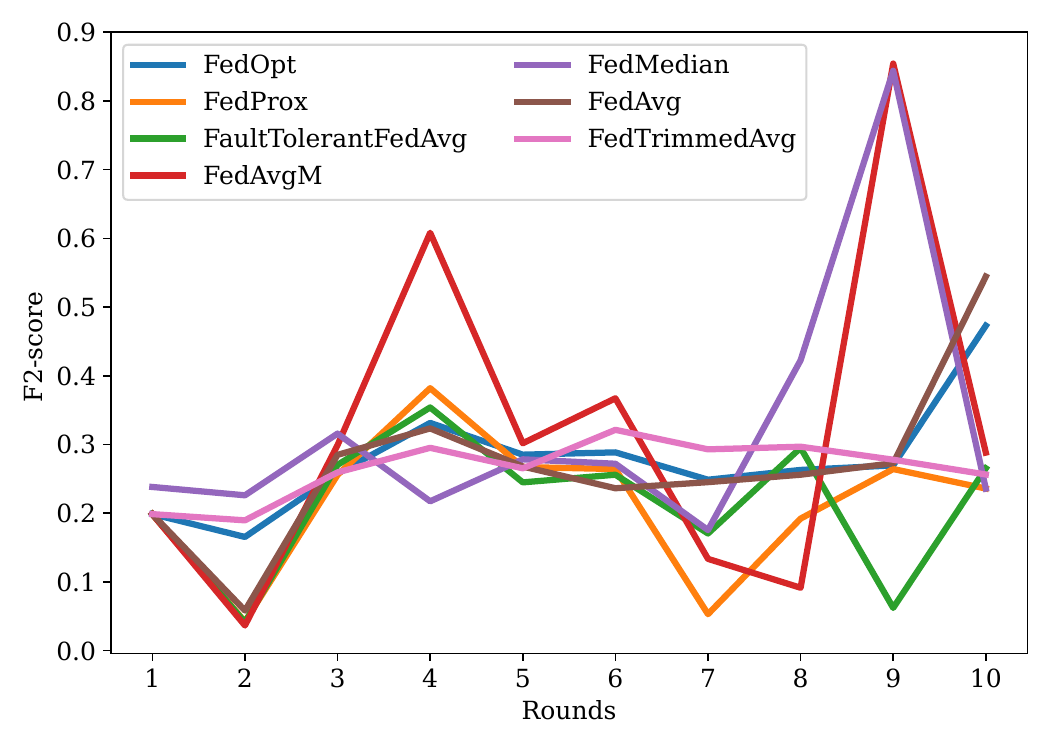}
        \caption{Evaluation results for all state-of-the-art FL methods exemplary on the real-world highway dataset.}
    \label{fig_AvgM_selection}
    \end{minipage}
\end{wrapfigure}

%\begin{figure}[!t]
%    \centering
%    \includegraphics[trim=10 10 10 10, clip, width=0.65\linewidth]{images/AvgM_selection.pdf}
%    \caption{Evaluation results for all state-of-the-art FL methods exemplary on the real-world highway dataset.}
%    \label{fig_AvgM_selection}
%\end{figure}

First, we discuss evaluation results on state-of-the-art methods to establish a baseline for the FL methods. Next, we present evaluation results on all three GNSS datasets and compare the MSE and MMD loss using confusion matrices. We analyze feature embeddings and evaluate the adaptation to different environmental scenarios in detail. All experiments are conducted utilizing Nvidia Tesla V100-SXM2 GPUs with 32 GB VRAM, equipped with Core Xeon CPUs and 192 GB RAM. We use the vanilla SGD optimizer with a learning rate set to $0.01$, and $0.05$ for the controlled environment; a decay rate of $5 \times 10^4$; a batch size of 40; and train each local model for 10 rounds with a maximum of 5 epochs per round. Throughout, we provide the accuracy achieved after round 10, the average accuracy standard deviation ($\pm$) results over 5 trainings (in \%) as well as the F1-score and F2-score. The best results are highlighted in \textbf{bold}. The best baseline result is \underline{underlined}.

\begin{table}[!t]
\begin{center}
    \caption{Evaluation results on the real-world highway dataset for the adaptation to new interference classes.}
    \label{table_evaluation_results_real_GNSS}
    \begin{tabular}{ p{0.5cm} | p{0.5cm} | p{0.5cm} | p{0.5cm} }
    \multicolumn{1}{c|}{\textbf{Method}} & \multicolumn{1}{c|}{\textbf{F1-score}} & \multicolumn{1}{c|}{\textbf{F2-score}} & \multicolumn{1}{c}{\textbf{Accuracy}} \\ \hline
    \multicolumn{1}{l|}{FedAvgM\cite{FedAvgM}} & \multicolumn{1}{r|}{0.10\,$\pm$\,0.05} & \multicolumn{1}{r|}{\underline{0.28}\,$\pm$\,0.13} & \multicolumn{1}{r}{\underline{98.31}\,$\pm$\,0.08} \\
    \multicolumn{1}{l|}{FedMedian\cite{FedMedian}} & \multicolumn{1}{r|}{\underline{0.13}\,$\pm$\,0.01} & \multicolumn{1}{r|}{0.26\,$\pm$\,0.03} & \multicolumn{1}{r}{98.06\,$\pm$\,0.00} \\
    \multicolumn{1}{l|}{FedTrim.Avg\cite{FedtrimmedAvg}} & \multicolumn{1}{r|}{0.08\,$\pm$\,0.04} & \multicolumn{1}{r|}{0.17\,$\pm$\,0.08} & \multicolumn{1}{r}{93.78\,$\pm$\,0.06} \\
    \multicolumn{1}{l|}{FedAvg\cite{Fedavg}} & \multicolumn{1}{r|}{0.06\,$\pm$\,0.05} & \multicolumn{1}{r|}{0.13\,$\pm$\,0.10} & \multicolumn{1}{r}{89.91\,$\pm$\,0.07} \\
    \multicolumn{1}{l|}{FedProx\cite{FedProx}} & \multicolumn{1}{r|}{0.07\,$\pm$\,0.04} & \multicolumn{1}{r|}{0.15\,$\pm$\,0.08} & \multicolumn{1}{r}{94.15\,$\pm$\,0.04} \\
    \multicolumn{1}{l|}{FedOpt\cite{FedOpt}} & \multicolumn{1}{r|}{0.09\,$\pm$\,0.05} & \multicolumn{1}{r|}{0.19\,$\pm$\,0.10} & \multicolumn{1}{r}{93.26\,$\pm$\,0.07} \\
    \multicolumn{1}{l|}{FaultTol.FedAvg\cite{faulttolerant}} & \multicolumn{1}{r|}{0.01\,$\pm$\,0.00} & \multicolumn{1}{r|}{0.03\,$\pm$\,0.01} & \multicolumn{1}{r}{78.18\,$\pm$\,0.09} \\
    \multicolumn{1}{l|}{FL (MSE)} & \multicolumn{1}{r|}{0.36\,$\pm$\,0.28} & \multicolumn{1}{r|}{0.49\,$\pm$\,0.31} & \multicolumn{1}{r}{93.31\,$\pm$\,0.12} \\
    \multicolumn{1}{l|}{FL (MMD)} & \multicolumn{1}{r|}{\textbf{0.46}\,$\pm$\,0.19} & \multicolumn{1}{r|}{\textbf{0.64}\,$\pm$\,0.18} & \multicolumn{1}{r}{\textbf{99.51}\,$\pm$\,0.00} \\
    \end{tabular}
\end{center}
\end{table}

\begin{table}[!t]
\begin{center}
    \caption{Evaluation results on the controlled GNSS small-scale dataset for the adaptation to new interference classes.}
    \label{table_evaluation_results_controlled_GNSS}
    \begin{tabular}{ p{0.5cm} | p{0.5cm} | p{0.5cm} | p{0.5cm} }
    \multicolumn{1}{c|}{\textbf{Method}} & \multicolumn{1}{c|}{\textbf{F1-score}} & \multicolumn{1}{c|}{\textbf{F2-score}} & \multicolumn{1}{c}{\textbf{Accuracy}} \\ \hline
    \multicolumn{1}{l|}{FedAvgM\cite{FedAvgM}} & \multicolumn{1}{r|}{0.96\,$\pm$\,0.05} & \multicolumn{1}{r|}{0.96\,$\pm$\,0.05} & \multicolumn{1}{r}{95.61\,$\pm$\,0.05} \\
    \multicolumn{1}{l|}{FedMedian\cite{FedAvgM}} & \multicolumn{1}{r|}{0.96\,$\pm$\,0.00} & \multicolumn{1}{r|}{0.97\,$\pm$\,0.00} & \multicolumn{1}{r}{95.81\,$\pm$\,0.00} \\
    \multicolumn{1}{l|}{FedTrimmedAvg\cite{FedtrimmedAvg}} & \multicolumn{1}{r|}{0.95\,$\pm$\,0.02} & \multicolumn{1}{r|}{0.97\,$\pm$\,0.01} & \multicolumn{1}{r}{95.08\,$\pm$\,0.03} \\
    \multicolumn{1}{l|}{FedAvg\cite{Fedavg}} & \multicolumn{1}{r|}{0.96\,$\pm$\,0.02} & \multicolumn{1}{r|}{0.97\,$\pm$\,0.01} & \multicolumn{1}{r}{96.17\,$\pm$\,0.02} \\
    \multicolumn{1}{l|}{FedProx\cite{FedProx}} & \multicolumn{1}{r|}{\underline{0.98}\,$\pm$\,0.00} & \multicolumn{1}{r|}{\underline{0.98}\,$\pm$\,0.00} & \multicolumn{1}{r}{\underline{97.98}\,$\pm$\,0.00} \\
    \multicolumn{1}{l|}{FedOpt\cite{FedOpt}} & \multicolumn{1}{r|}{0.96\,$\pm$\,0.01} & \multicolumn{1}{r|}{0.97\,$\pm$\,0.01} & \multicolumn{1}{r}{95.59\,$\pm$\,0.01} \\
    \multicolumn{1}{l|}{FaultTol.FedAvg\cite{faulttolerant}} & \multicolumn{1}{r|}{0.96\,$\pm$\,0.04} & \multicolumn{1}{r|}{0.97\,$\pm$\,0.03} & \multicolumn{1}{r}{96.52\,$\pm$\,0.00} \\
    \multicolumn{1}{l|}{Our method (MSE)} & \multicolumn{1}{r|}{0.98\,$\pm$\,0.00} & \multicolumn{1}{r|}{0.97\,$\pm$\,0.01} & \multicolumn{1}{r}{97.94\,$\pm$\,0.00} \\
    \multicolumn{1}{l|}{Our method (MMD)} & \multicolumn{1}{r|}{\textbf{0.98}\,$\pm$\,0.00} & \multicolumn{1}{r|}{\textbf{0.98}\,$\pm$\,0.00} & \multicolumn{1}{r}{\textbf{98.17}\,$\pm$\,0.00} \\
    \end{tabular}
\end{center}
\end{table}

\begin{table}[!t]
\begin{center}
    \caption{Evaluation results on the controlled GNSS large-scale dataset for the adaptation to new environments.}
    \label{table_evaluation_results_envs_GNSS}
    \begin{tabular}{ p{0.5cm} | p{0.5cm} | p{0.5cm} | p{0.5cm} }
    \multicolumn{1}{c|}{\textbf{Method}} & \multicolumn{1}{c|}{\textbf{F1-score}} & \multicolumn{1}{c|}{\textbf{F2-score}} & \multicolumn{1}{c}{\textbf{Accuracy}} \\ \hline
    \multicolumn{1}{l|}{FedAvgM\cite{FedAvgM}} & \multicolumn{1}{r|}{\underline{\textbf{0.9999}}} & \multicolumn{1}{r|}{0.9997} & \multicolumn{1}{r}{99.973} \\
    \multicolumn{1}{l|}{FedMedian\cite{FedMedian}} & \multicolumn{1}{r|}{0.9998} & \multicolumn{1}{r|}{0.9997} & \multicolumn{1}{r}{99.979} \\
    \multicolumn{1}{l|}{FedTrimmedAvg\cite{FedtrimmedAvg}} & \multicolumn{1}{r|}{0.9998} & \multicolumn{1}{r|}{0.9996} & \multicolumn{1}{r}{99.970} \\
    \multicolumn{1}{l|}{FedAvg\cite{Fedavg}} & \multicolumn{1}{r|}{0.9998} & \multicolumn{1}{r|}{0.9997} & \multicolumn{1}{r}{99.979} \\
    \multicolumn{1}{l|}{FedProx\cite{FedProx}} & \multicolumn{1}{r|}{0.9997} & \multicolumn{1}{r|}{0.9996} & \multicolumn{1}{r}{99.966} \\
    \multicolumn{1}{l|}{FedOpt\cite{FedOpt}} & \multicolumn{1}{r|}{0.9998} & \multicolumn{1}{r|}{0.9997} & \multicolumn{1}{r}{99.974} \\
    \multicolumn{1}{l|}{FaultTol.FedAvg\cite{faulttolerant}} & \multicolumn{1}{r|}{\underline{0.9998}} & \multicolumn{1}{r|}{0.9998} & \multicolumn{1}{r}{\underline{99.981}}\\
    \multicolumn{1}{l|}{Our method (MSE)} & \multicolumn{1}{r|}{ 0.9998} & \multicolumn{1}{r|}{0.9997} & \multicolumn{1}{r}{99.977} \\
    \multicolumn{1}{l|}{Our method (MMD)} & \multicolumn{1}{r|}{\textbf{0.9999}} & \multicolumn{1}{r|}{\textbf{ 0.9998}} & \multicolumn{1}{r}{\textbf{99.989}} \\
    \end{tabular}
\end{center}
\end{table}

\textbf{State-of-the-art Methods.} Figure~\ref{fig_AvgM_selection} presents the initial results of seven state-of-the-art methods~\cite{FedAvgM,FedMedian,FedtrimmedAvg,Fedavg,FedProx,FedOpt,faulttolerant} applied to the real-world highway dataset. Among these methods, FedAvgM demonstrates the highest and most consistent F2-score, particularly from rounds 3 to 6. This observation prompted the incorporation of the FedAvgM aggregation step into our FL method. Moreover, our MMD-based early stopping method can be integrated into any FL method. Interestingly, none of the methods exhibit stability across all rounds. This trend stems from the inherent complexity posed by novel classes encountered at various local stations, which often share similarities with previously encountered classes. However, our method effectively addresses this challenge by dynamically adjusting epochs for each local station.

\textbf{Evaluation Results.} Table~\ref{table_evaluation_results_real_GNSS} and \ref{table_evaluation_results_controlled_GNSS} present the evaluation results of all baseline methods, as well as our early stopping-based FL method (using MSE and MMD). On the real-world highway dataset, FedAvgM achieved the highest accuracy at 98.31\%. By utilizing our MMD-based early stopping technique, we improved the average accuracy to 99.51\%. Additionally, the F1 (0.46) and F2-score (0.64) were significantly increased by reducing the number of false negatives. With the MSE loss, the accuracy was lower (93.31\%), indicating a domain shift between data and underscoring the importance of selecting an optimal discrepancy measure. On the controlled small-scale dataset, FedProx achieved the highest baseline results with 97.98\%. Our proposed techniques increased this to 97.94\% with the MSE loss and 98.17\% with the MMD loss, with equal F1-/F2-scores compared to FedProx. On both datasets, we reached the upper bound by specifically training the model on the complete dataset.

\textbf{Adaptation to Scenarios.} Table~\ref{table_evaluation_results_envs_GNSS} presents the results of adapting to new environments on the controlled large-scale dataset. Although there are substantial differences between environmental scenarios due to multipath effects, which result in decreased accuracy, class-incremental learning on this dataset is feasible. Consequently, the differences among all methods are marginal. Nonetheless, our MMD-based technique outperforms all other methods, achieving an accuracy of 99.989\%. It is important to note, however, that there are several samples that are not unique to each scenario.

\begin{figure}[!t]
    \centering
	\begin{minipage}[t]{0.493\linewidth}
        \centering
    	\includegraphics[trim=7 32 44 70, clip, width=1.0\linewidth]{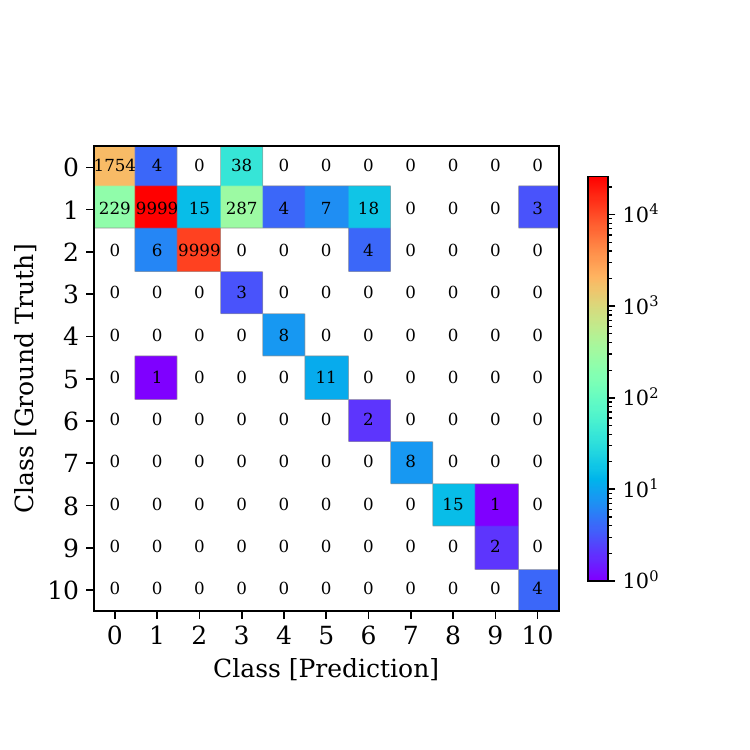}
        \subcaption{MSE loss.}
        \label{fig_cm_real_world1}
    \end{minipage}
    \hfill
	\begin{minipage}[t]{0.493\linewidth}
        \centering
    	\includegraphics[trim=7 32 44 70, clip, width=1.0\linewidth]{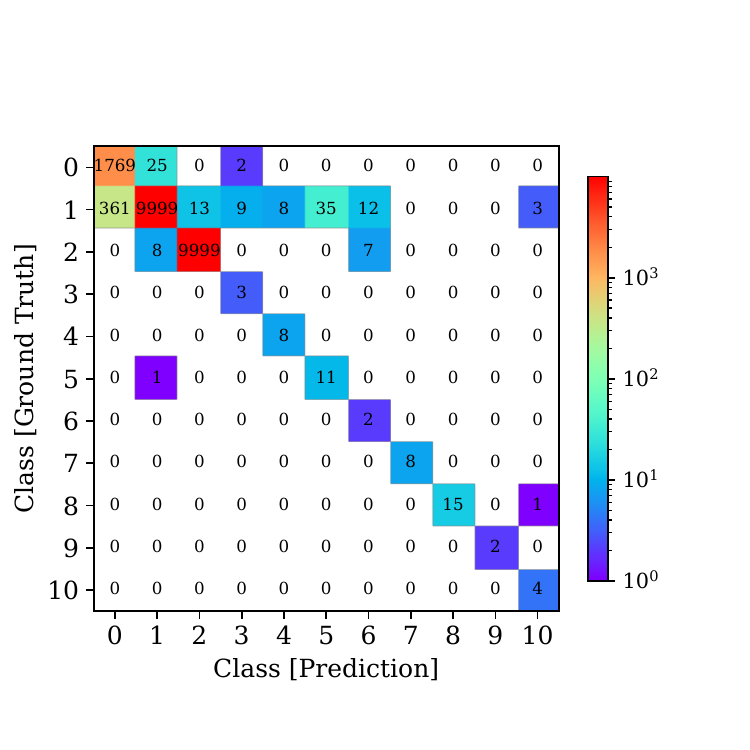}
        \subcaption{MMD loss.}
        \label{fig_cm_real_world2}
    \end{minipage}
    \caption{Confusion matrices for the adaptation to new classes on the real-world GNSS dataset.}
    \label{fig_cm_real_world}
\end{figure}

\begin{figure}[!t]
    \centering
	\begin{minipage}[t]{0.493\linewidth}
        \centering
    	\includegraphics[trim=7 32 44 70, clip, width=1.0\linewidth]{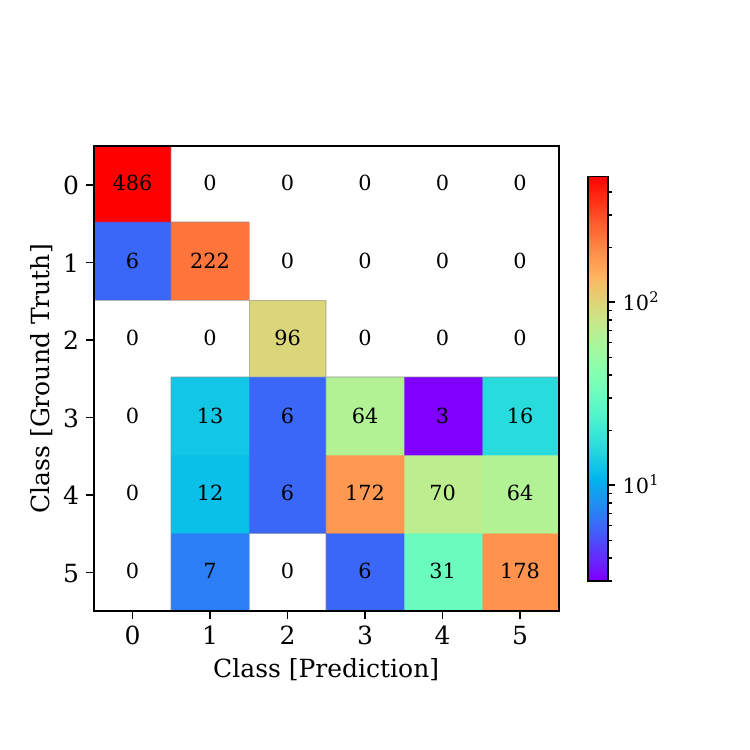}
        \subcaption{MSE loss.}
        \label{fig_cm_small_scale1}
    \end{minipage}
    \hfill
	\begin{minipage}[t]{0.493\linewidth}
        \centering
    	\includegraphics[trim=7 32 44 70, clip, width=1.0\linewidth]{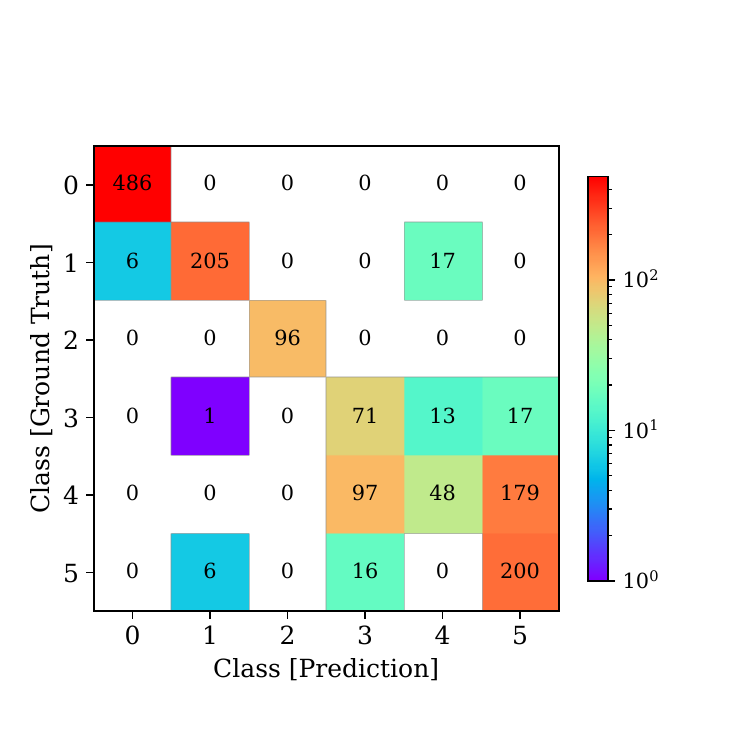}
        \subcaption{MMD loss.}
        \label{fig_cm_small_scale2}
    \end{minipage}
    \caption{Confusion matrices for the adaptation to new classes on the controlled small-scale dataset.}
    \label{fig_cm_small_scale}
\end{figure}

\begin{figure}[!t]
    \centering
	\begin{minipage}[t]{0.493\linewidth}
        \centering
    	\includegraphics[trim=7 32 44 70, clip, width=1.0\linewidth]{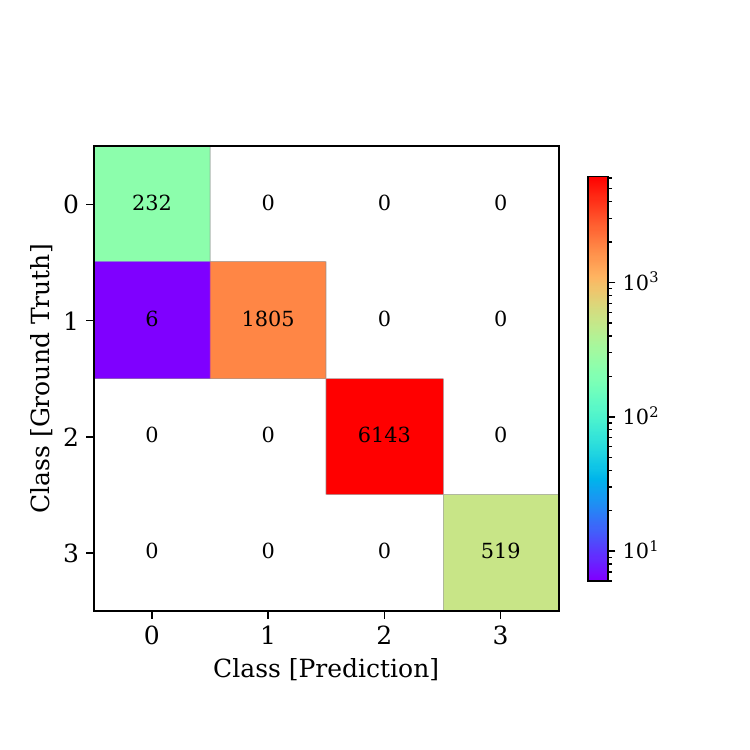}
        \subcaption{MSE loss.}
        \label{fig_cm_large_scale_env1}
    \end{minipage}
    \hfill
	\begin{minipage}[t]{0.493\linewidth}
        \centering
    	\includegraphics[trim=7 32 44 70, clip, width=1.0\linewidth]{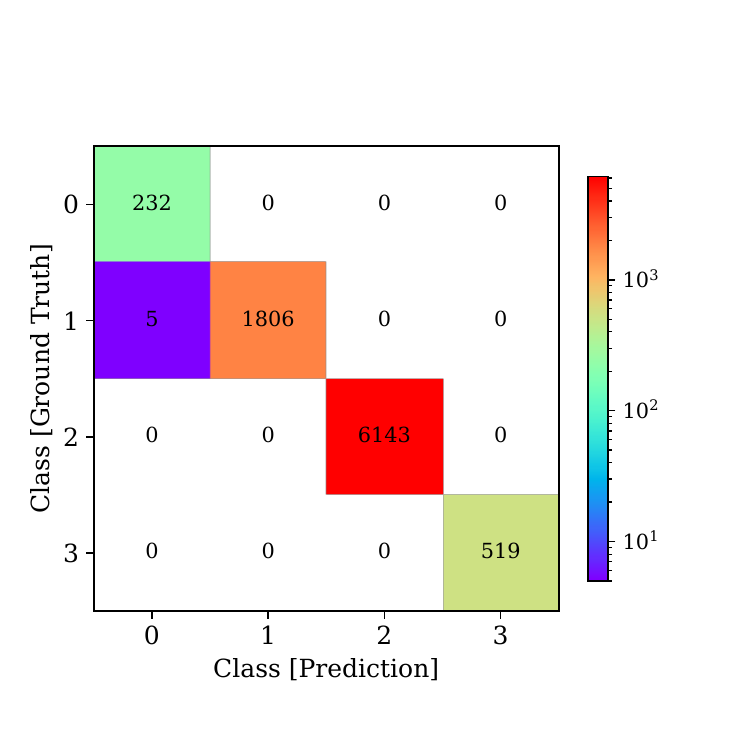}
        \subcaption{MMD loss.}
        \label{fig_cm_large_scale_env2}
    \end{minipage}
    \caption{Confusion matrices for the adaptation to new scenarios on the controlled large-scale dataset.}
    \label{fig_cm_large_scale_env}
\end{figure}

\begin{figure}[!t]
    \centering
	\begin{minipage}[t]{0.325\linewidth}
        \centering
    	\includegraphics[trim=7 24 25 70, clip, width=1.0\linewidth]{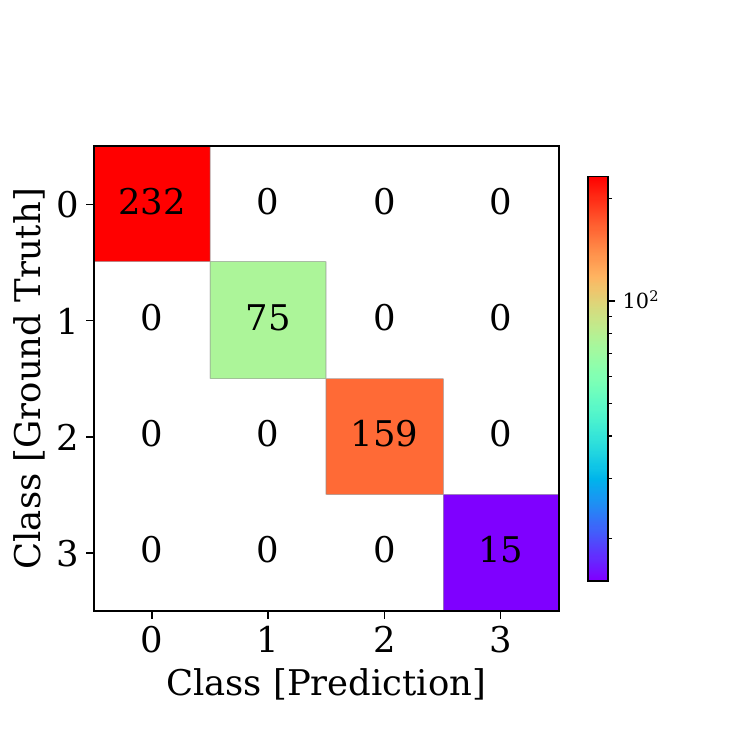}
        \subcaption{Scenario 1.}
        \label{figdifferent_scenario_ablation_study1}
    \end{minipage}
    \hfill
	\begin{minipage}[t]{0.325\linewidth}
        \centering
    	\includegraphics[trim=7 24 25 70, clip, width=1.0\linewidth]{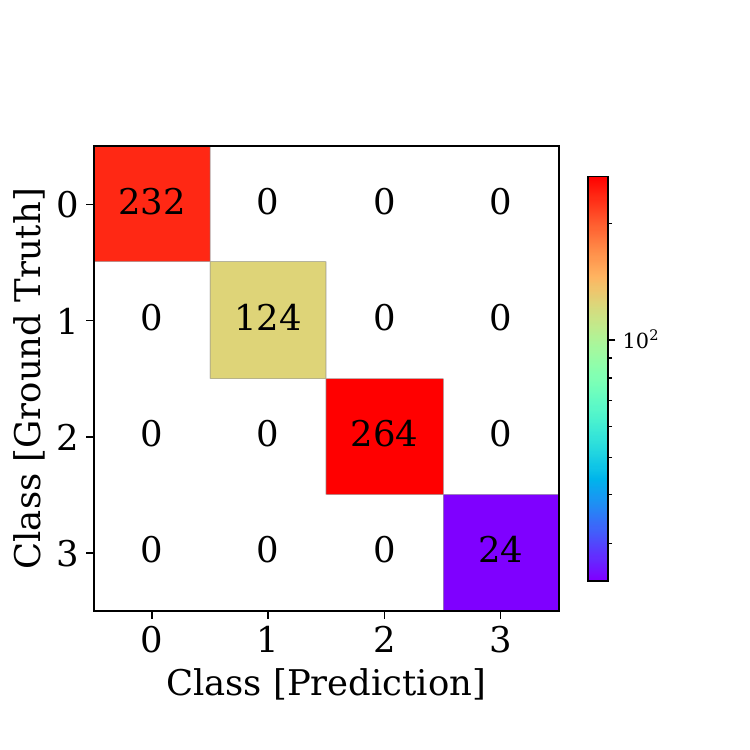}
        \subcaption{Scenario 6.}
        \label{figdifferent_scenario_ablation_study2}
    \end{minipage}
    \hfill
	\begin{minipage}[t]{0.325\linewidth}
        \centering
    	\includegraphics[trim=7 24 25 70, clip, width=1.0\linewidth]{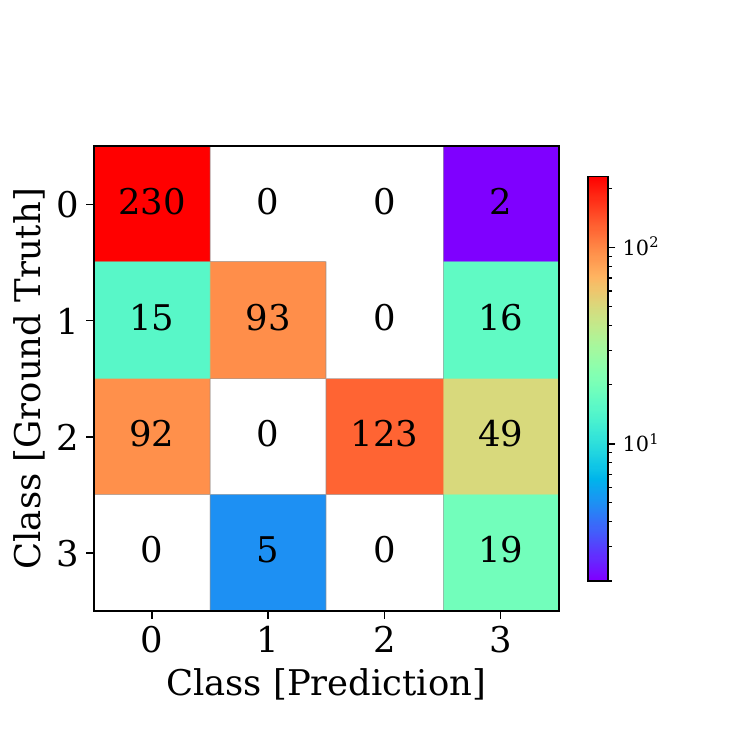}
        \subcaption{Scenario 8.}
        \label{figdifferent_scenario_ablation_study3}
    \end{minipage}
	\begin{minipage}[t]{0.325\linewidth}
        \centering
    	\includegraphics[trim=7 24 25 70, clip, width=1.0\linewidth]{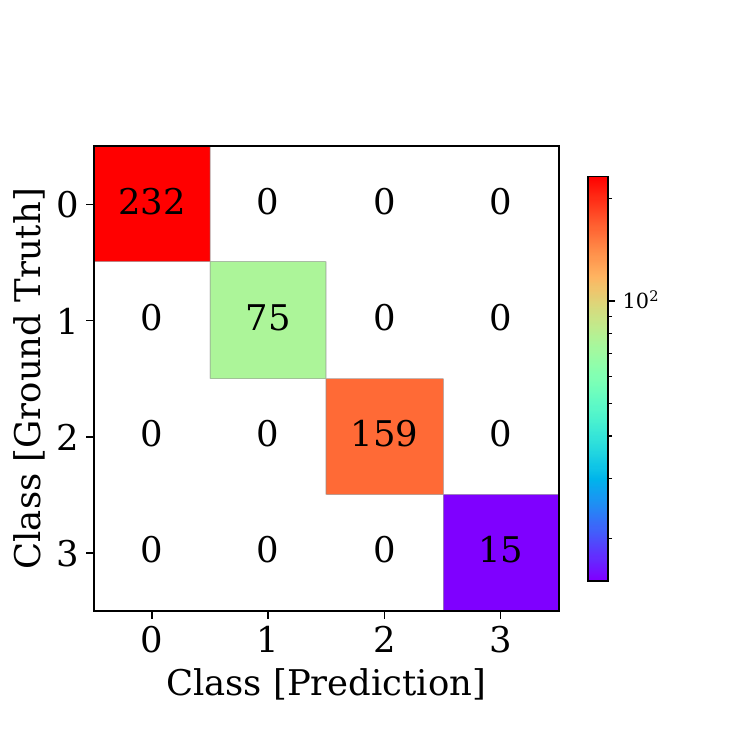}
        \subcaption{Scenario 1.}
        \label{figdifferent_scenario_ablation_study4}
    \end{minipage}
    \hfill
	\begin{minipage}[t]{0.325\linewidth}
        \centering
    	\includegraphics[trim=7 24 25 70, clip, width=1.0\linewidth]{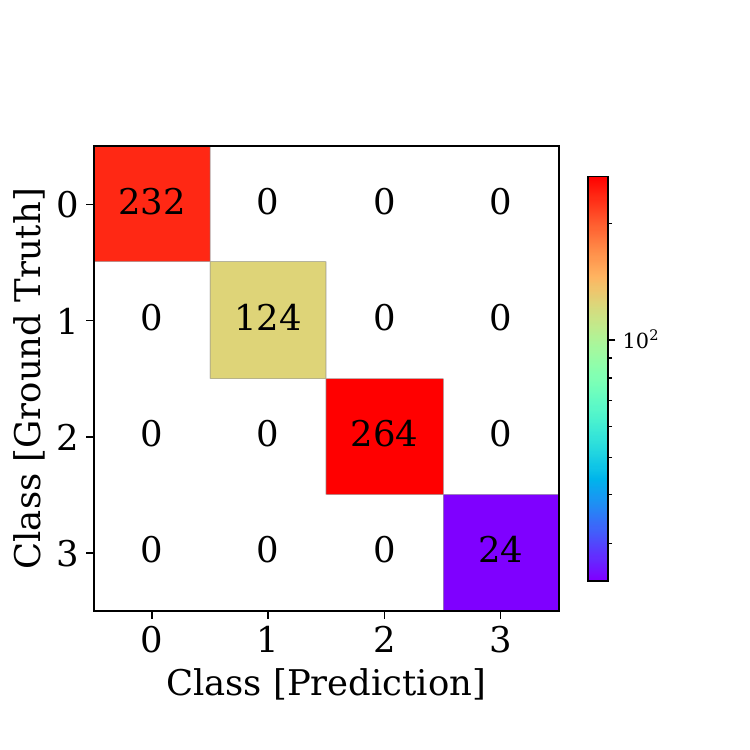}
        \subcaption{Scenario 6.}
        \label{figdifferent_scenario_ablation_study5}
    \end{minipage}
    \hfill
	\begin{minipage}[t]{0.325\linewidth}
        \centering
    	\includegraphics[trim=7 24 25 70, clip, width=1.0\linewidth]{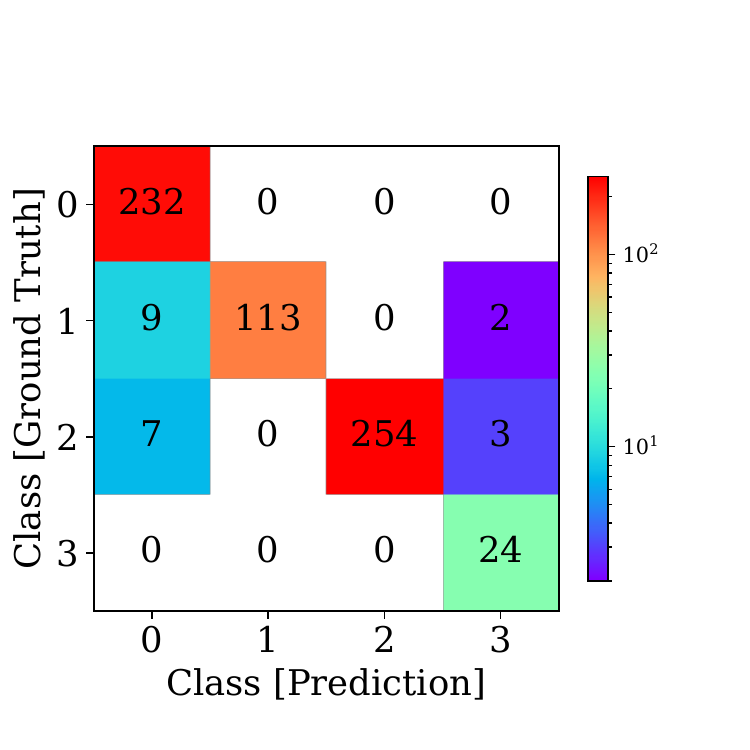}
        \subcaption{Scenario 8.}
        \label{figdifferent_scenario_ablation_study6}
    \end{minipage}
    \caption{Results for FL on scenarios 1, 3, and 6 (top row) and the scenarios 1, 6, and 8 (bottom row) on the controlled large-scale dataset. Sub-captions show the test datasets.}
    \label{figdifferent_scenario_ablation_study}
\end{figure}

\textbf{Confusion Matrices.} Next, we analyze the MSE and MMD loss functions alongside the confusion matrices for all datasets. Figure~\ref{fig_cm_real_world} presents the results for the real-world dataset, illustrating that the high number of false negative samples (class 1 of Figure~\ref{fig_cm_real_world1}) can be significantly reduced. By utilizing the MMD loss (Figure~\ref{fig_cm_real_world2}), the discrepancy between classes can be optimally evaluated and the number of epochs can be dynamically adjusted. Conversely, the MMD-based early stopping method reduces false positives (e.g., for classes 4 and 5) in the controlled small-scale dataset (Figure~\ref{fig_cm_small_scale}). Aligned with the results in Table~\ref{table_evaluation_results_envs_GNSS}, adaptation to the new environments is achievable with high accuracy (refer to Figure~\ref{fig_cm_large_scale_env}). For the controlled large-scale dataset, we initially adapt to classes 1, 3, and 6, and test on classes 1, 6, and 8 (refer to Figures~\ref{figdifferent_scenario_ablation_study1} to \ref{figdifferent_scenario_ablation_study3}). This shows that adaptation to scenarios 1 and 6 is possible with 100\% accuracy, while performance for scenario 8 significantly drops due to different multipath effects. By adapting to scenario 8 instead of scenario 3 (refer to Figures~\ref{figdifferent_scenario_ablation_study4} to \ref{figdifferent_scenario_ablation_study6}), the model demonstrates high robustness. In summary, our models are not only capable of incrementally learning new classes with FL but also of adapting effectively to changes in the environment.

\setlength{\intextsep}{6pt}
\setlength{\columnsep}{12pt}
\begin{wrapfigure}{R}{4.6cm}
    \begin{minipage}[b]{1.0\linewidth}
        \includegraphics[trim=21 4 4 38, clip, width=1.0\linewidth]{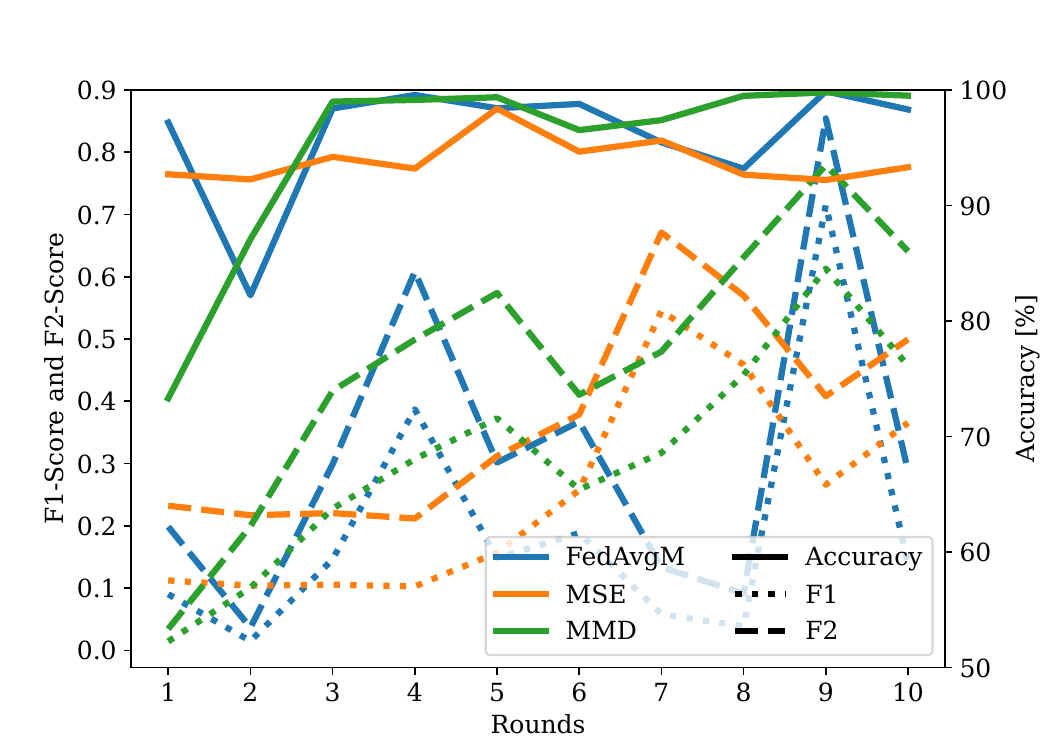}
        \caption{Comparison of FedAvgM and our embedding-based (MSE and MMD) FL method on the real-world GNSS dataset.}
        \label{fig_Ablation_study}
    \end{minipage}
\end{wrapfigure}

%\begin{figure}[!t]
%    \centering
%    \includegraphics[trim=21 4 4 38, clip, width=0.65\linewidth]{images/Abl_study.pdf}
%    \caption{Comparison of FedAvgM~\cite{FedAvgM} and our embedding-based (MSE and MMD) FL method on the real-world GNSS dataset.}
%    \label{fig_Ablation_study}
%\end{figure}

\textbf{Influence of MMD.} A more detailed evaluation of FedAvgM and our early stopping-based FL method on the real-world dataset is provided in Figure~\ref{fig_Ablation_study}. We present the average F1-score, F2-score, and accuracy for all 10 rounds, averaged over 5 training runs. While the accuracy of FedAvgM drops in rounds 7 and 8, our MMD-based technique remains robust and achieves the highest performance. However, the MSE-based technique, though robust across all rounds, exhibits lower classification accuracy. The high number of false negative samples results in significantly lower F1-scores and F2-scores. In contrast, the MMD-based method achieves an F2-score of 0.64 after the 10$^{\text{th}}$ round. We emphasize that our FL technique, which incorporates early stopping, trains faster compared to FedAvgM while maintaining greater robustness.

\begin{figure}[!t]
\centering
    \begin{subfigure}{0.323\linewidth}
      \centering
      \includegraphics[trim=53 39 50 42, clip, width=1.0\linewidth]{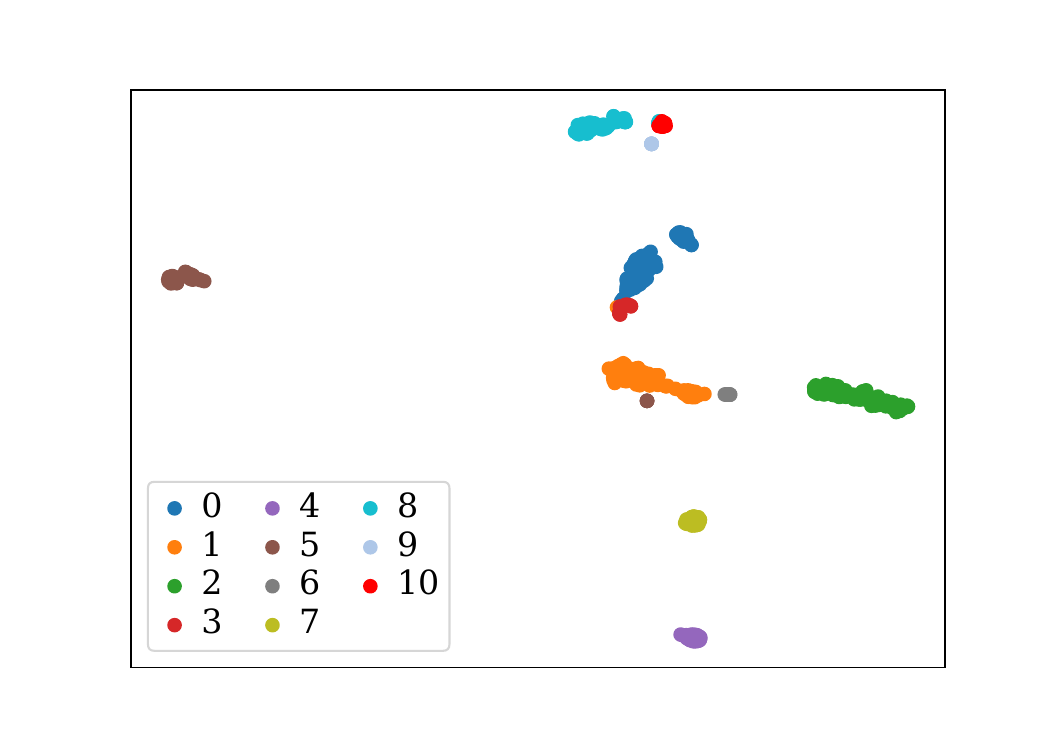}
      \caption{FedAvgM.}
      \label{fig_embedding_comparisonl1}
    \end{subfigure}
    \hfill
    \begin{subfigure}{0.323\linewidth}
      \centering
      \includegraphics[trim=53 39 50 42, clip, width=1.0\linewidth]{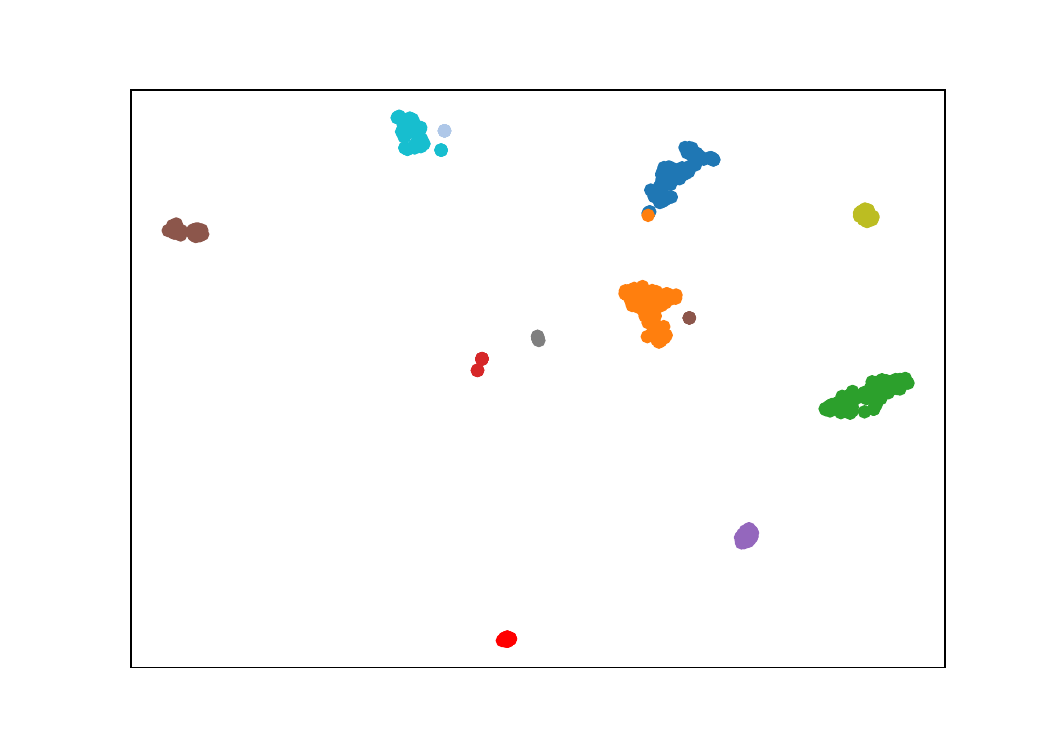}
      \caption{FL (MSE loss).}
      \label{fig_embedding_comparisonl2}
    \end{subfigure}
    \hfill
    \begin{subfigure}{0.323\linewidth}
      \centering
      \includegraphics[trim=53 39 50 42, clip, width=1.0\linewidth]{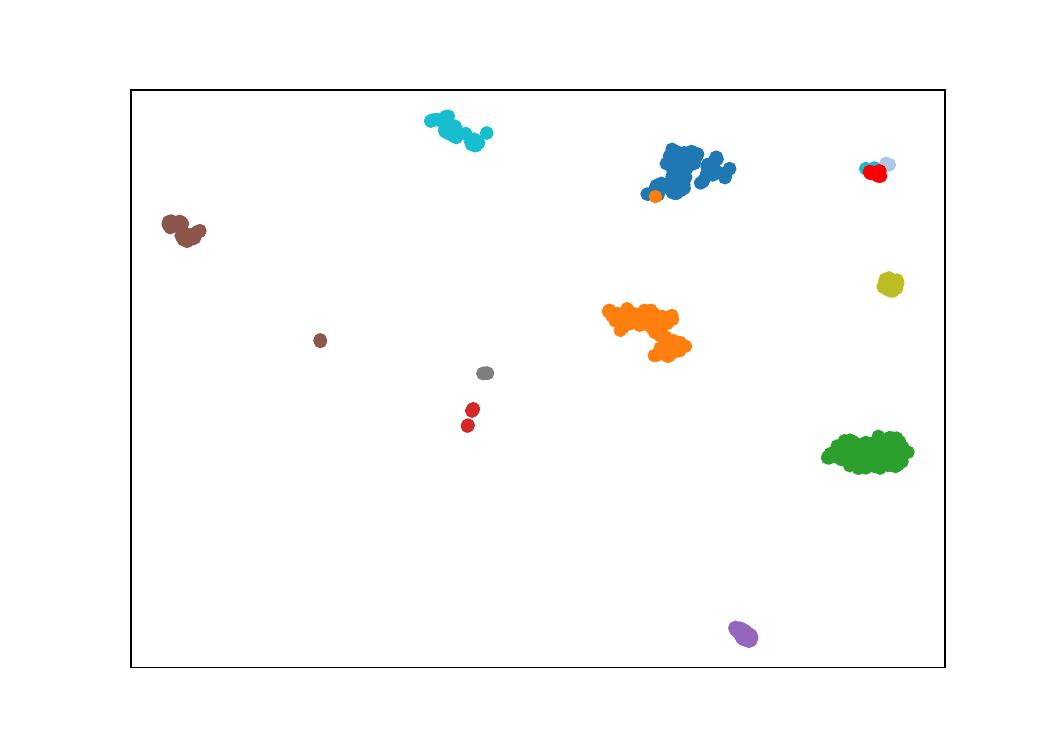}
      \caption{FL (MMD loss).}
      \label{fig_embedding_comparisonl3}
    \end{subfigure}
    \caption{Comparison of t-SNE embeddings for all 11 classes on the real-world GNSS highway dataset.}
    \label{fig_embedding_comparisonl}
\end{figure}

\textbf{Feature Embedding Analysis.} Figure~\ref{fig_embedding_comparisonl} illustrates the feature embedding of the last convolutional layer, which has an output size of 512, utilizing the t-distributed stochastic neighbor embedding (t-SNE)~\cite{maaten_hinton} with a perplexity of 30, an initial momentum of 0.5, and a final momentum of 0.8. Exemplifying the real-world highway dataset, we incorporate all 11 classes for FedAvgM (Figure~\ref{fig_embedding_comparisonl1}) and our early stopping-based FL method with the MSE loss (Figure~\ref{fig_embedding_comparisonl2}) and the MMD loss (Figure~\ref{fig_embedding_comparisonl3}). For FedAvgM, most classes are grouped on the right side of the embedding, whereas for our FL-based method, the classes are better separated. Specifically, there is no confusion between classes 1 and 5 with the MMD loss, although there is a different grouping between classes 8, 9, and 10. Notably, the MMD method significantly increases the inter-class distances compared to the MSE approach. This enhanced separability suggests that the MMD method facilitates better discrimination between classes, potentially leading to improved classification performance.

\begin{figure}[!t]
    \centering
	\begin{minipage}[t]{0.48\linewidth}
        \centering
    	\includegraphics[trim=10 27 22 52, clip, width=0.78\linewidth]{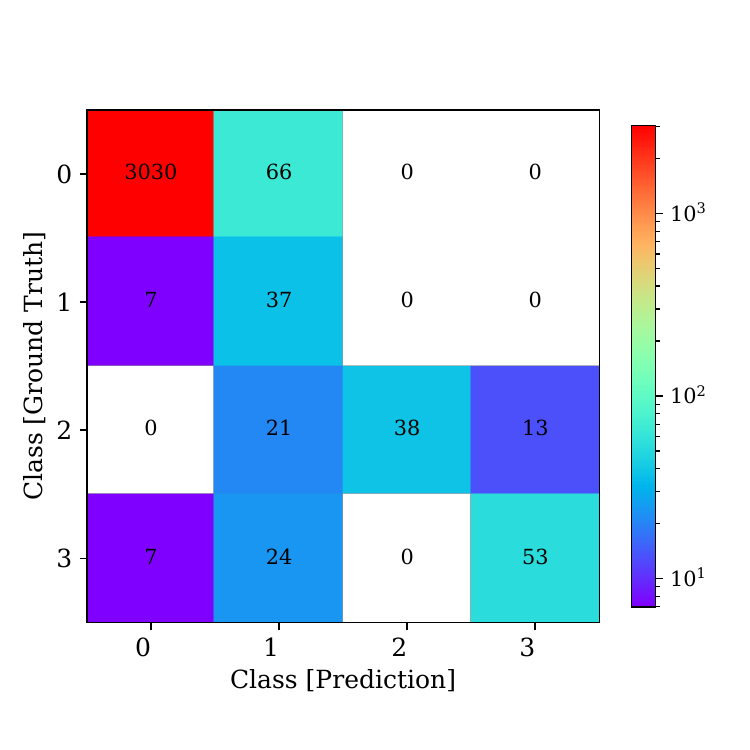}
        \caption{Confusion matrix for adapting to new interference classes between highway 1 and highway 2.}
        \label{figure_highway_confusion_matrix}
    \end{minipage}
    \hfill
	\begin{minipage}[t]{0.493\linewidth}
        \centering
    	\includegraphics[trim=14 72 38 84, clip, width=1.0\linewidth]{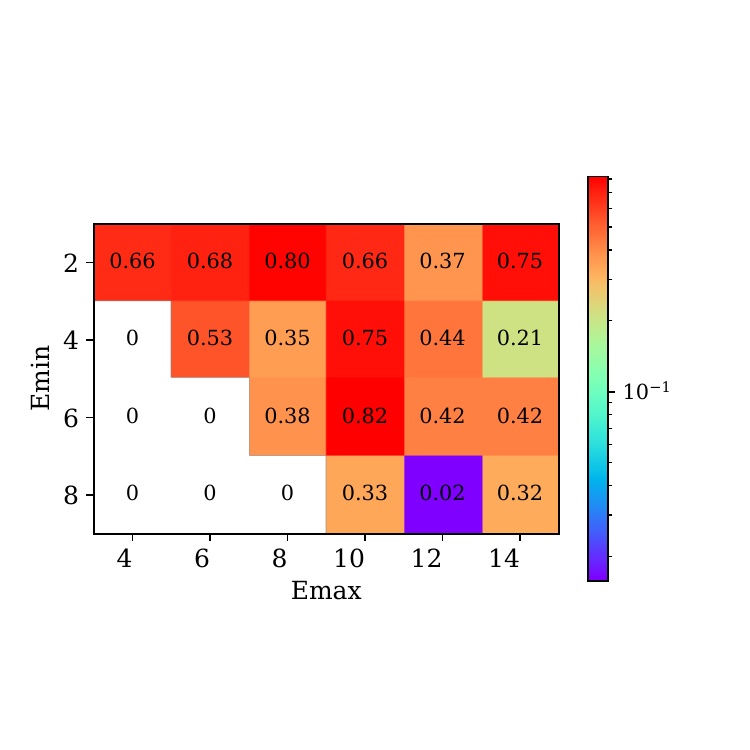}
        \caption{Evaluation (F2-score) for various $E_{\text{min}}$ and $E_{\text{max}}$ parameters on the real-world highway dataset.}
        \label{fig_e_min_e_max}
    \end{minipage}
\end{figure}

\begin{figure}[!t]
    \centering
	\begin{minipage}[t]{0.493\linewidth}
        \centering
    	\includegraphics[trim=32 140 100 156, clip, width=1.0\linewidth]{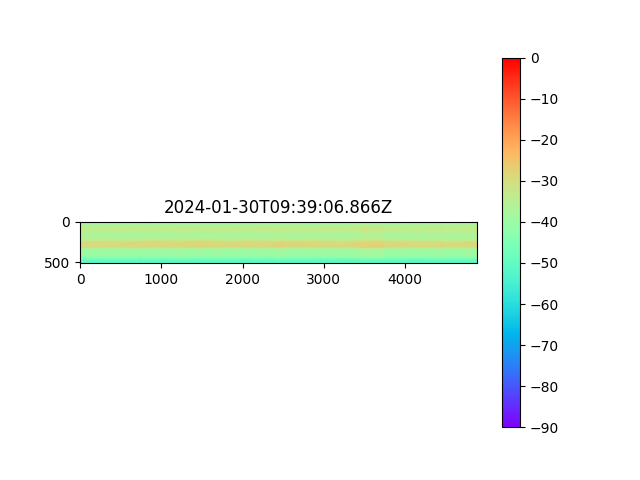}
        \subcaption{Class 1.}
        \label{fig_snapshots_example1}
    \end{minipage}
    \hfill
	\begin{minipage}[t]{0.493\linewidth}
        \centering
    	\includegraphics[trim=32 140 100 156, clip, width=1.0\linewidth]{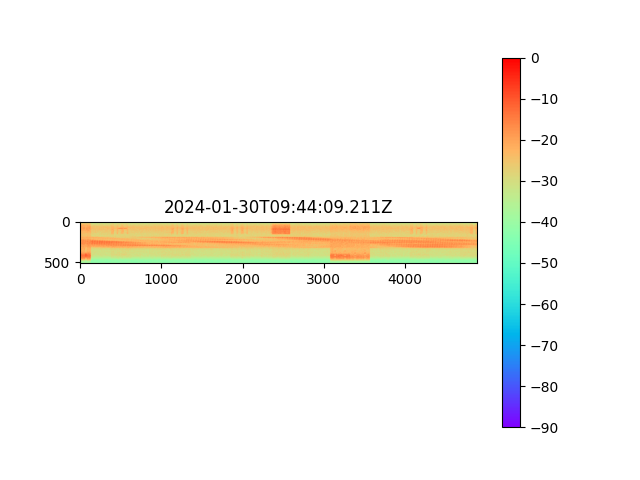}
        \subcaption{Class 1.}
        \label{fig_snapshots_example2}
    \end{minipage}
	\begin{minipage}[t]{0.493\linewidth}
        \centering
    	\includegraphics[trim=32 140 100 156, clip, width=1.0\linewidth]{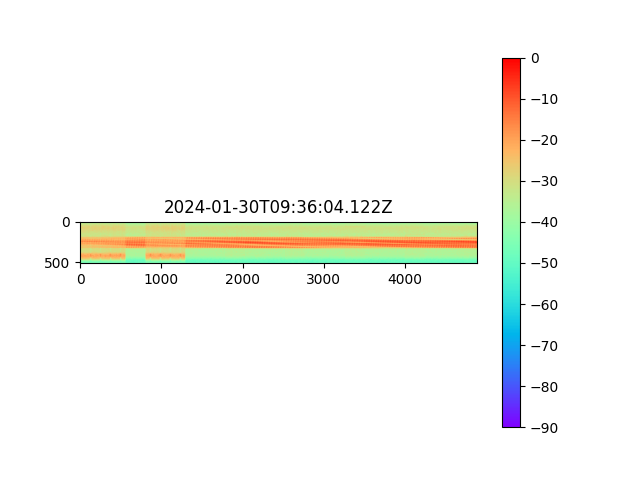}
        \subcaption{Class 2.}
        \label{fig_snapshots_example3}
    \end{minipage}
    \hfill
	\begin{minipage}[t]{0.493\linewidth}
        \centering
    	\includegraphics[trim=32 140 100 156, clip, width=1.0\linewidth]{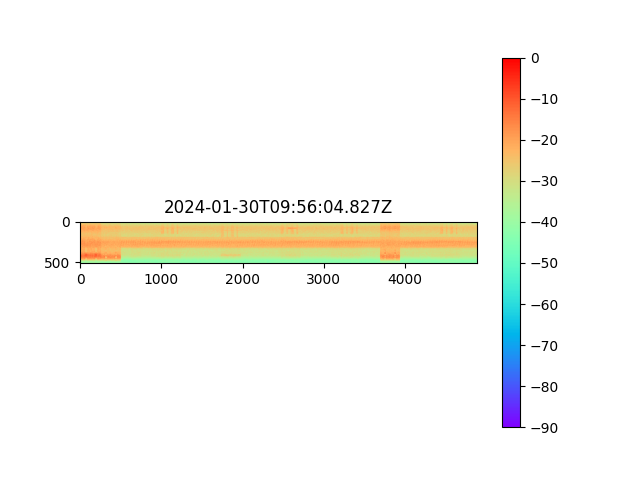}
        \subcaption{Class 3.}
        \label{fig_snapshots_example4}
    \end{minipage}
    \caption{Exemplary snapshots with difficult classification tasks.}
    \label{fig_snapshots_example}
\end{figure}

\textbf{Adaptation to Highway Environment.} In this section, we discuss the adaptation process from one highway to another for the orchestration of sensor stations. Figure~\ref{figure_highway_confusion_matrix} presents the evaluation results of this adaptation to three new interference classes. For class 1, the model demonstrates high classification accuracy. However, there is notable confusion between class 2 and classes 1 and 2, as well as between class 3 and classes 1 and 2. The final test accuracy is 95.81\%, with an F1-score of 0.823, an F2-score of 0.884, a precision of 0.738, and a recall of 0.930, indicating that the FL model maintains a high level of accuracy. Although the model was trained on three positive classes from the new dataset, it is still capable of predicting negative classes from the previous dataset. Figure~\ref{fig_snapshots_example} provides examples of challenging samples through illustrative snapshots. These 20\,ms snapshots were used to generate 1\,ms samples, all labeled with the same class. The snapshot for class 2 (Figure~\ref{fig_snapshots_example3}) contains random noise at the beginning, which results in incorrect classification, while the rest is correctly labeled as class 5, exhibiting significant interference. Similarly, the snapshot for class 3 (Figure~\ref{fig_snapshots_example4}) includes noisy regions. However, class 1 and class 3 share similarities due to the presence of similar jammer types, making it difficult to set a threshold (as seen in the varying intensities in class 1). In conclusion, robust adaptation would require more positively labeled samples.

\textbf{Ablation Study on Number of Epochs.} For smaller values of $E_{\text{min}}$, such as $E_{\text{min}} = 2$, higher values of $E_{\text{max}}$ result in significant performance improvements (refer to Figure~\ref{fig_e_min_e_max}). This indicates that stations with lower initial training requirements benefit more from extended training on more challenging tasks. Conversely, when $E_{\text{min}}$ is larger, e.g., $E_{\text{min}} = 6$ or $E_{\text{min}} = 8$, the performance gains from increasing $E_{\text{max}}$ are smaller or less consistent. This suggests that local models requiring more minimal training epochs already achieve a satisfactory level of performance, making further training less impactful. Additionally, performance decreases substantially for certain combinations of $E_{\text{min}}$ and $E_{\text{max}}$, implying that some models, particularly those with higher difficulty levels (higher $E_{\text{max}}$), may require more specialized training or adaptive techniques to improve results. This highlights a trade-off between training time and performance: while increasing $E_{\text{max}}$ generally enhances performance for easier tasks (lower $E_{\text{min}}$), the benefit diminishes as $E_{\text{min}}$ increases. Therefore, models with simpler tasks require fewer epochs to perform well, whereas more difficult models may need more intensive training to yield only marginal improvements.

%% file: 06conclusion.tex
\section{Summary}
\label{label_conclusion}

\subsection{Discussion}
\label{label_summary_discussion}

In this paper, we successfully demonstrated the adaptation to new interference classes in both real-world environments with highly imbalanced classes and controlled indoor settings. For the indoor scenarios, we employed two distinct datasets, differentiated by the use of low-frequency and high-frequency antennas for data acquisition, to facilitate the adaptation to new interference classes and scenarios characterized by varying degrees of multipath effects. Our FL approach achieved high classification accuracies (exceeding 98\%) across all three tasks and an accuracy of 95.81\% for adapting to novel classes at a second highway stations. Consequently, we established the feasibility of adapting to new sensor stations at different locations, which involves the combination of various adaptation tasks. However, data collection at multiple locations in the presence of real-world jammers poses significant challenges due to the illegal use of jamming devices~\cite{mountin,german_bill}. Additionally, the high cost of recording stations and the logistical difficulties of installing them near highway bridges further complicate this process~\cite{gentili_mirchandani,yang_yang,nugroho_vishnoi}. Our method assumes a majority of non-faulty nodes, but in high-failure or malicious environments, integrating fault-tolerant algorithms, Byzantine consensus protocols \cite{gao_xu_fan}, or anomaly detection \cite{xu_jiang_zeng_li} can enhance resilience by isolating faulty nodes and maintaining system integrity.

However, there may be debate regarding the necessity of using a federated approach, particularly in light of concerns about confidential or sensitive data, as a centralized training method could be an alternative. Many modern applications, deployed across various devices, rely on geographical information and, consequently, on geolocation data. The definition of personal data encompasses all information that relates to an identified or identifiable natural person. This includes device IDs, location data, browser types, IP addresses, and similar identifiers. Even if the data controller does not possess or have access to the identifying information, the data may still be classified as personal if another entity could identify the individual using the available information -- this applies to GNSS data as well~\cite{lueghausen_lachenmann}. For instance, GNSS-based activity tracking data is a valuable resource for understanding active modes of transportation. However, such data also raises privacy concerns, as movement recordings of individuals are considered sensitive~\cite{brauer_maekinen}. Additionally, Gonz'{a}lez et al.~\cite{gonzalez_lay} identified emerging threats to participant confidentiality and anonymity, issues with \textit{unanticipated} data collection and exploitation, difficulties in obtaining properly informed consent, and concerns about the representation of vulnerable populations who have limited access to smartphones and face legitimate fears of surveillance. In conclusion, we opted for a federated approach.

\subsection{Conclusion}
\label{label_summary_con}

We presented a FL method incorporating an early stopping approach to balance underrepresented class labels and focus on challenging classes. Specifically, we compute the number of epochs per FL round based on the discrepancy between feature embeddings, utilizing the maximum mean discrepancy (MMD). The MMD loss demonstrates superior performance compared to the standard mean squared error loss. On three GNSS datasets for interference classification, our method achieves an accuracy of 99.51\% in adapting to new classes on a real-world dataset, 98.17\% in adapting to new classes in a controlled small-scale environment, and 99.989\% in adapting to new scenarios in a controlled large-scale environment. Our proposed technique surpasses all state-of-the-art FL methods. Our experiments demonstrated the feasibility of orchestrating sensor stations for GNSS interference monitoring with the capability to adapt to new classes.